\documentclass{article}

\usepackage[final]{cpal_2026}

\usepackage{url}
\usepackage{hyperref}

\usepackage{xcolor}
\usepackage{amsmath}
\usepackage{amssymb}
\usepackage{amsthm}
\usepackage[ruled, lined, linesnumbered, commentsnumbered, longend, noend]{algorithm2e}
\usepackage{booktabs}
\usepackage{multirow}
\usepackage{subcaption}
\usepackage{graphicx}
\usepackage{colortbl}
\usepackage{cleveref}

\title{Learning in the Null Space: Small Singular Values for Continual Learning}

\author{
  Cuong Anh Pham\textsuperscript{1}, ~Praneeth Vepakomma\textsuperscript{1, 2}, ~Samuel Horv\'ath\textsuperscript{1} \\
  \textsuperscript{1}Mohamed bin Zayed University of Artificial Intelligence (MBZUAI), UAE\\ \textsuperscript{2}Massachusetts Institute of Technology (MIT), USA\\
  \texttt{\{cuong.pham, praneeth.vepakomma, samuel.horvath\}@mbzuai.ac.ae}
}

\newcommand{\W}{\mathbf{W}}

\begin{document}

\maketitle

\begin{abstract}
Alleviating catastrophic forgetting while enabling further learning is a primary challenge in continual learning (CL). Orthogonal-based training methods have gained attention for their efficiency and strong theoretical properties, and many existing approaches enforce orthogonality through gradient projection.
In this paper, we revisit orthogonality and exploit the fact that small singular values correspond to directions that are nearly orthogonal to the input space of previous tasks. Building on this principle, we introduce NESS (Null-space Estimated from Small Singular values), a CL method that applies orthogonality directly in the weight space rather than through gradient manipulation. Specifically, NESS constructs an approximate null space using the smallest singular values of each layer’s input representation and parameterizes task-specific updates via a compact low-rank adaptation (LoRA-style) formulation constrained to this subspace. The subspace basis is fixed to preserve the null-space constraint, and only a single trainable matrix is learned for each task. This design ensures that the resulting updates remain approximately in the null space of previous inputs while enabling adaptation to new tasks.
Our theoretical analysis and experiments on three benchmark datasets demonstrate competitive performance, low forgetting, and stable accuracy across tasks, highlighting the role of small singular values in continual learning.
The code is available at \href{https://github.com/pacman-ctm/NESS}{\texttt{https://github.com/pacman-ctm/NESS}}.
  
\end{abstract}

\section{Introduction}

Continual learning (CL) is a machine learning approach that trains a single model through a sequence of tasks with varying properties while maintaining performance on all previous tasks. In CL, \textit{catastrophic forgetting} is the most significant challenge, in which the performance of learned tasks decreases drastically after continual training on new tasks \citep{wang2024comprehensive}. Addressing this phenomenon is the key to designing an effective CL algorithm. 
As machine learning models are deployed in more complex, dynamic data and applications, the need for a stable model that reduces forgetting is emerging as a key characteristic for designing an efficient CL algorithm. Due to the need to adapt to new tasks, CL approaches must find a balance between \textit{plasticity} (the ability to adapt to new information) and \textit{stability} (the ability to retain past knowledge) \citep{coleman2025parameter}. 

There are many approaches to building an efficient CL model with a low forgetting rate, and Singular Value Decomposition (SVD)-based methods have been a popular choice for maintaining performance after training on new tasks. Early approaches to mitigating forgetting focused primarily on regularizing the weight changes~\citep{rebuffi2017icarl, shin2017continual, rolnick2019experience}. More recent approaches, on the other hand, were based on storing episodic memories from previous data and adding constraints based on this information to the gradient direction, thus reducing the forgetting of old tasks~\citep{lopez2017gradient, chaudhryefficient}.

\begin{figure}[ht!]
    \centering
    \includegraphics[width=0.95\linewidth]{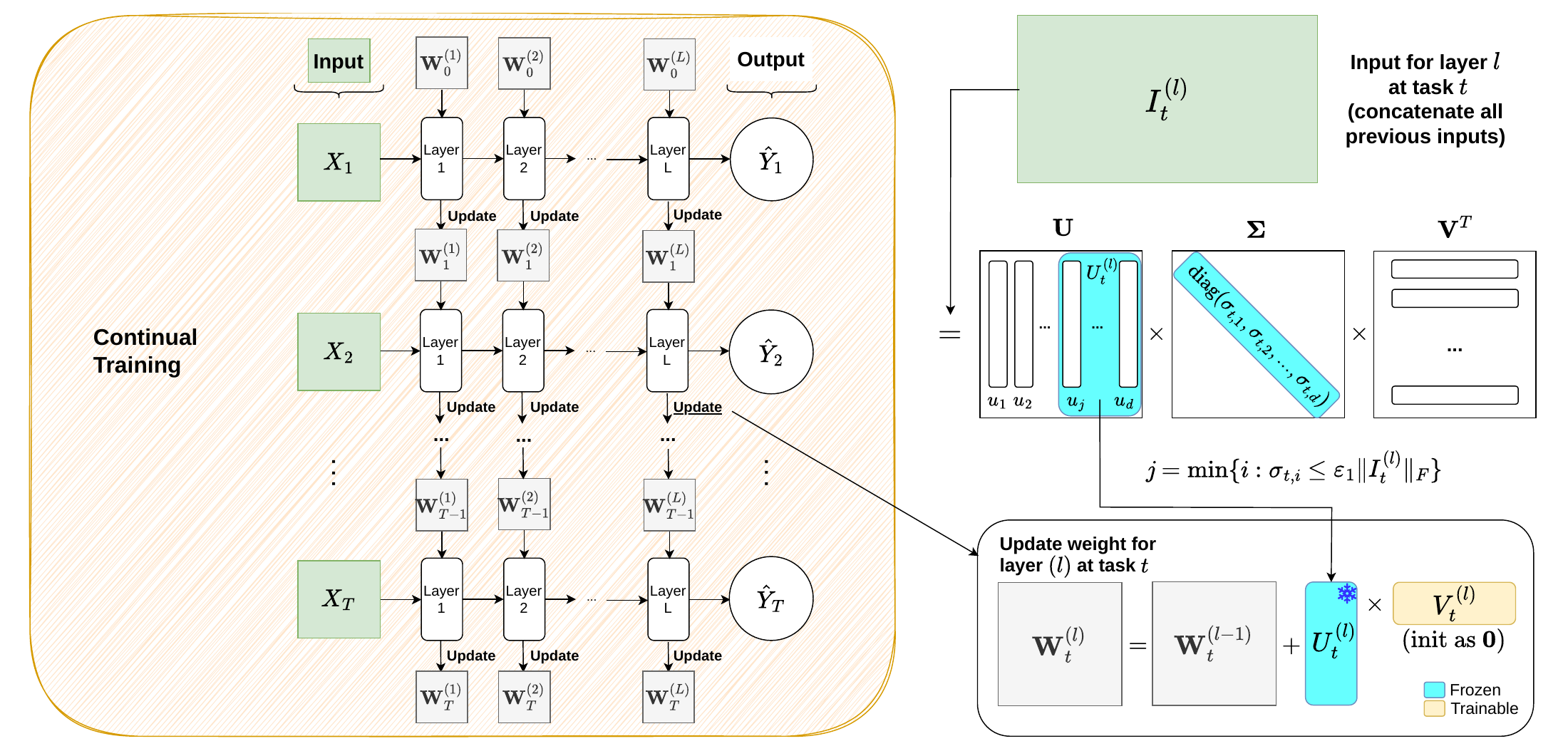}
    \caption{Overview of NESS. During continual training (left), each task updates the network sequentially. For every task $t$ and layer $l$, we collect the concatenated inputs from previous tasks and compute the SVD of the corresponding covariance matrix (right). The update is parameterized as a low-rank decomposition $\Delta W_t^{(l)} = U_t^{(l)} V_t^{(l)}$, where $U_t^{(l)}$ is a frozen orthogonal basis constructed from singular vectors associated with small singular values, and $V_t^{(l)}$ is a trainable matrix initialized to zero. This structured update constrains learning to an approximate null subspace of previous inputs, limiting interference while enabling adaptation to the current task.}
    \label{fig:our_method}
\end{figure}

Using gradient projection is currently favored for mitigating catastrophic forgetting, with the pioneering work GPM proposed by \citet{saha2021gradient}, which utilized the null space of previously learned tasks for gradient projection, thus preventing interference with learned knowledge. They also proposed a \textit{backward transfer} metric to measure forgetting during continual training, which is widely used to evaluate later CL models. 

A large body of SVD-based continual learning methods relies on a common mechanism: enforcing orthogonality through gradient projection. In practice, these approaches compute the SVD of input representations from previous tasks, identify the dominant subspace associated with the largest singular values, and project gradients for subsequent tasks onto its orthogonal complement. In this way, updates are constrained to avoid directions that could interfere with previously learned representations.

Importantly, projecting gradients onto the complement of the dominant subspace is equivalent to directly operating in the subspace spanned by the singular vectors corresponding to the smallest singular values. Both viewpoints impose the same geometric constraint. The difference lies in how the constraint is implemented: prior work enforces it during optimization via gradient projection.

In contrast, we incorporate this constraint directly into the parameterization. Rather than modifying gradients at each step, we construct a fixed orthogonal basis from the singular vectors associated with small singular values and use it to define the allowable update space (see Figure~\ref{fig:our_method}). For each layer, this basis remains fixed, and only a single trainable matrix is learned within this restricted space. As a result, the orthogonality condition is satisfied by construction, and learning proceeds directly within the approximate null space of previous inputs regardless of the selected optimizer.

\begin{figure}[t]
    \centering
    \begin{subfigure}[b]{0.48\textwidth}
        \centering
        \includegraphics[width=\textwidth]{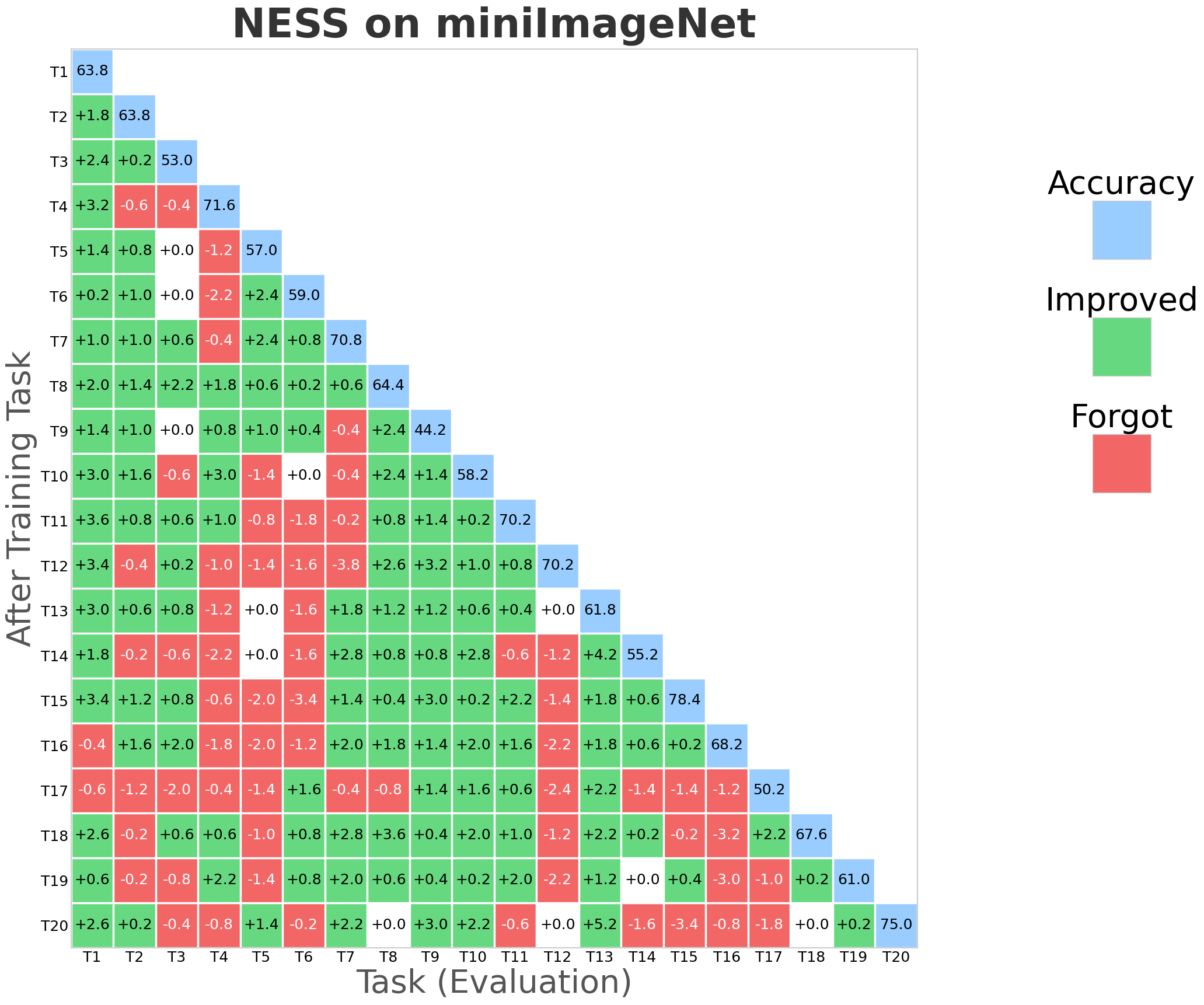}
        \caption{NESS}
    \end{subfigure}
    \hfill
    \begin{subfigure}[b]{0.48\textwidth}
        \centering
        \includegraphics[width=\textwidth]{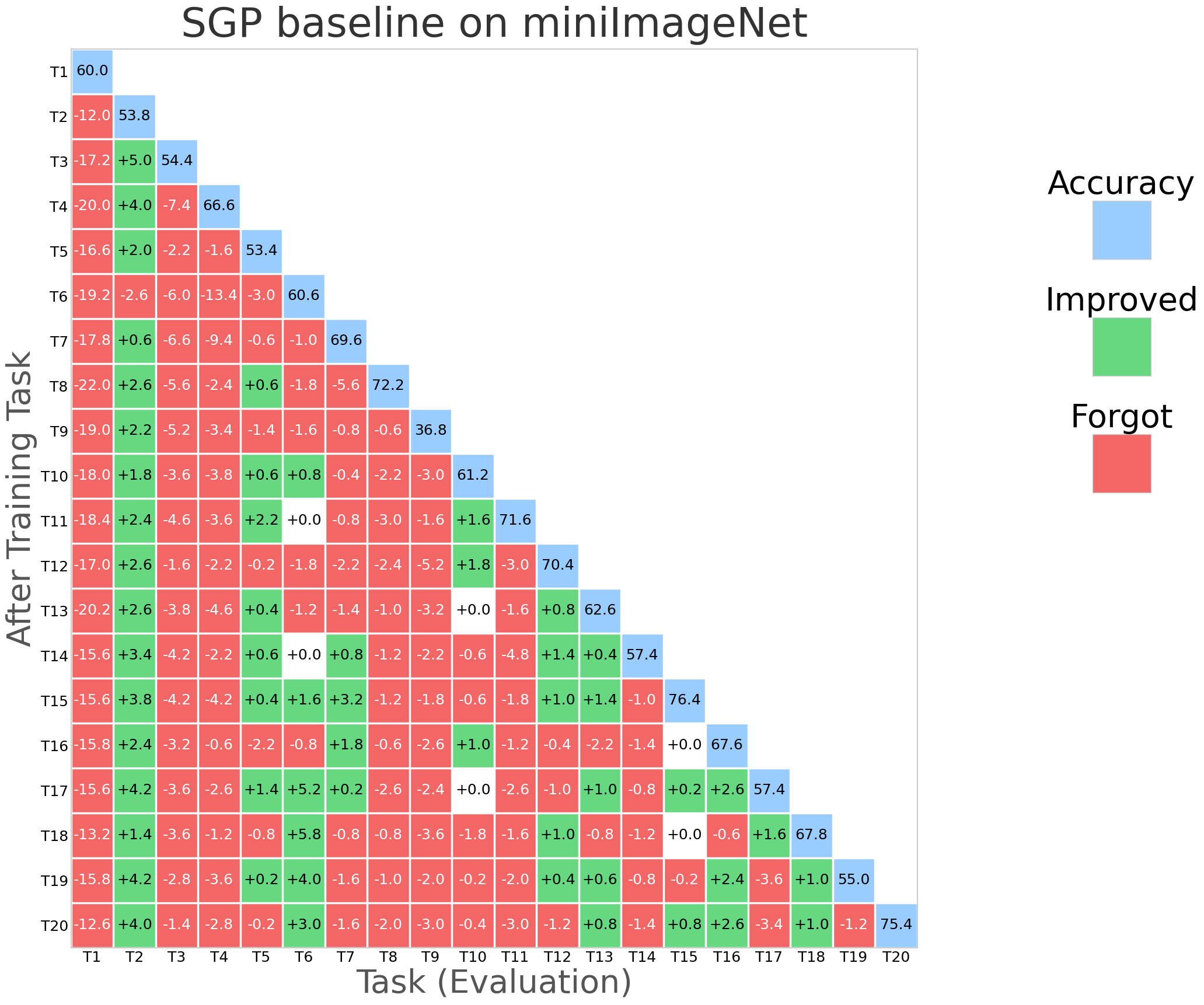}
        \caption{SGP baseline}
    \end{subfigure}
    
    \vspace{0.5cm}
    
    \begin{subfigure}[b]{0.48\textwidth}
        \centering
        \includegraphics[width=\textwidth]{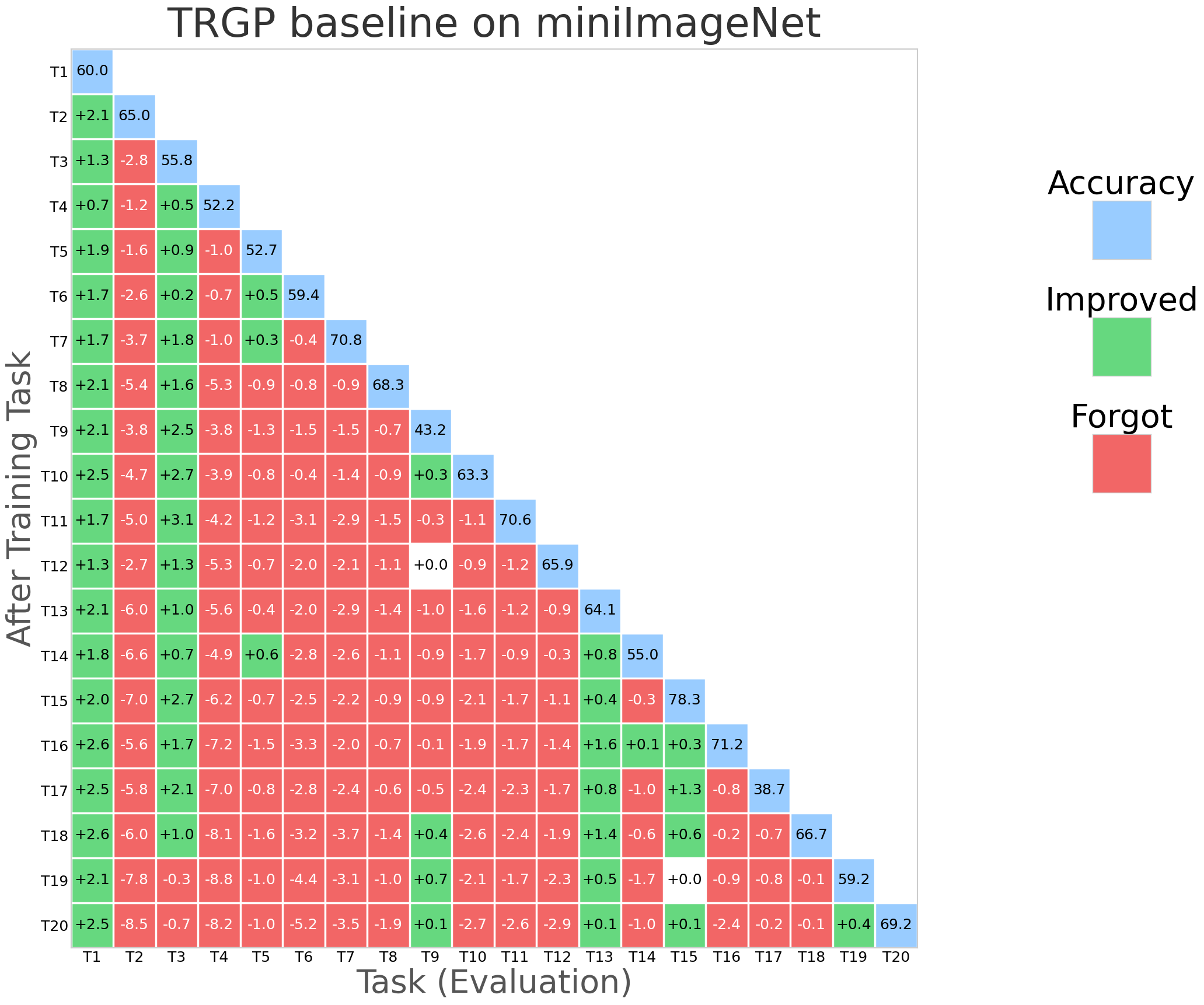}
        \caption{TRGP baseline}
    \end{subfigure}
    \hfill
    \begin{subfigure}[b]{0.48\textwidth}
        \centering
        \includegraphics[width=\textwidth]{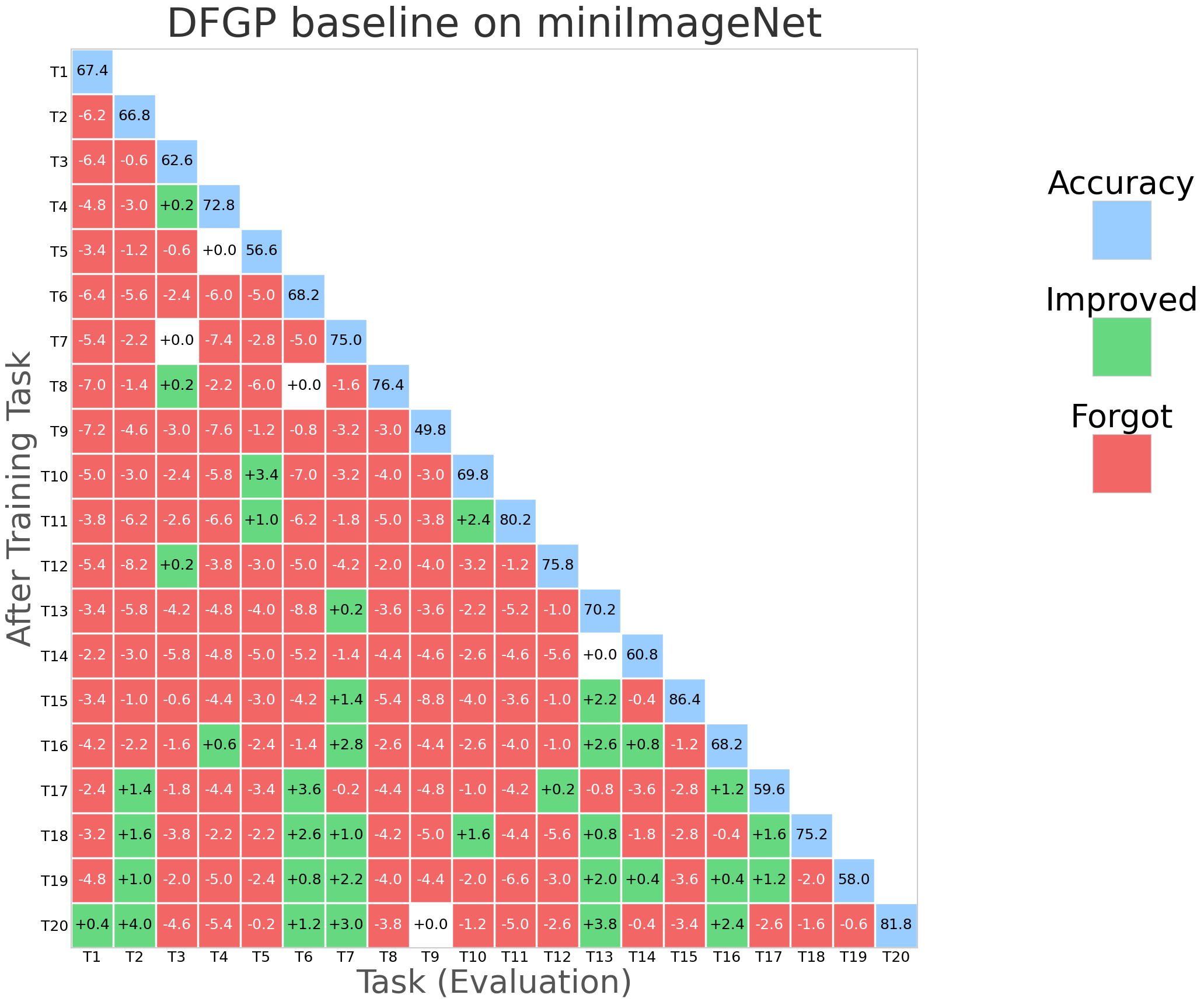}
        \caption{DFGP baseline}
    \end{subfigure}
    \caption{Performance (best setting of each method, focusing on forgetting rate, with green color denotes improving performance and red color denotes losing performance, compared to the blue color cell for the performance when the task first trained) of (a) NESS compared to (b) SGP; (c) TRGP; (d) DFGP on miniImageNet experiment (seed 3).}
    \label{fig:mini_seed3}
\end{figure}

We refer to this approach as \textbf{NESS (Null-space Estimated from Small Singular values)}. Empirical evaluations demonstrate that this parameterization leads to stable behavior across tasks while maintaining competitive performance (see Figure~\ref{fig:mini_seed3}).

Below is a summary of our contributions:
\begin{itemize}
    \item We propose \textbf{NESS}, a continual learning (CL) algorithm that enforces orthogonality by directly parameterizing weight updates within the approximate null space of previously learned feature spaces. This constraint is implemented through a fixed subspace basis derived from small singular values, while learning a single trainable matrix per layer. We provide a theoretical analysis supporting the efficiency of this formulation.
    \item We evaluate NESS against several orthogonality-based CL baselines on standard image classification benchmarks. The results demonstrate competitive performance and consistently improved backward transfer (i.e., reduced forgetting), indicating that NESS provides a stable and efficient approach to continual learning.
\end{itemize}

\section{Related Work}

\textbf{Catastrophic Forgetting in Continual Learning (CL).} 
Numerous approaches have been proposed to address catastrophic forgetting in continual learning \citep{van2024continual, wang2024comprehensive}. Existing methods are commonly categorized based on their underlying mechanisms.

\textit{Memory-based methods} maintain a buffer of samples from previous tasks and replay them while learning new tasks; a representative example is A-GEM \citep{chaudhryefficient}.  
\textit{Architecture-based methods} modify or expand the network structure during training to reduce interference across tasks, with early work such as HAT \citep{serra2018overcoming}.  
\textit{Optimization-based methods} introduce additional constraints or regularization terms into the loss function to preserve prior knowledge. These methods are often divided into regularization-based approaches (e.g., EWC \citep{kirkpatrick2017overcoming}) and subspace-based approaches \citep{wang2024comprehensive}.  
\textit{Representation-based methods} aim to learn task-invariant or structured representations that remain stable throughout the continual learning process.  
Finally, \textit{Bayesian methods} formulate continual learning from a probabilistic perspective, treating catastrophic forgetting as an inference problem.

\noindent\textbf{Orthogonality-based Continual Learning.}
Orthogonality-based approaches, which fall under optimization-based methods, constrain gradients to reduce interference across tasks. These methods can be broadly divided into two categories:

(i) \textit{Orthogonal gradient-based methods}, which enforce new task gradients to be orthogonal to the gradient subspace of previous tasks. This typically requires storing or estimating gradient information from earlier tasks.

(ii) \textit{Orthogonal feature-based methods}, which constrain gradients to be orthogonal to the subspace spanned by input features or representations of previous tasks. These approaches rely on estimating feature subspaces and generally offer improved scalability compared to gradient-storage methods.

\medskip
\noindent\textbf{Representative Orthogonality-based Approaches.}
Early works include OGD \citep{farajtabar2020orthogonal}, which restricts gradient directions to avoid interference, and OWM \citep{zeng2019continual}, which leverages a context-dependent processing module to reuse feature representations while updating the model.

Building on these ideas, \citet{saha2021gradient} proposed Gradient Projection Memory (GPM), which projects gradients onto the null space of previous task subspaces to prevent interference. Subsequent extensions improved different aspects of this framework. 
FS-DGPM \citep{deng2021flattening} incorporates sharpness-aware flattening to reduce the generalization gap. 
TRGP \citep{lin2022trgp} introduces a trust-region strategy to select important previous tasks and scales weight projections accordingly. 
SGP \citep{saha2023continual} improves scalability by combining null-space projection with scaled updates along retained gradient subspaces.

More recent developments include DFGP \citep{Yang_2023_ICCV, yang_2025_tpami}, which integrates mixup-based data augmentation \citep{zhang2018mixup, zhang2021does} and sharpness-aware minimization (SAM) \citep{foretsharpness} into the gradient projection framework to improve generalization. 
\citet{zhao2023rethinking} proposed Space Decoupling (SD), which decomposes the input space into complementary subspaces to balance stability and plasticity. 
Finally, CODE-CL \citep{Apolinario_2025_ICCV} introduces conceptor-based gradient projection, using conceptor matrices to project gradients onto pseudo-orthogonal subspaces of previous task input spaces.

While these methods enforce orthogonality through gradient-based constraints during optimization, we next formalize the continual learning setting and introduce an alternative parameterization that incorporates the orthogonality constraint directly in the weight space.

Recent work has also highlighted the importance of small-magnitude parameters in neural networks. 
\citet{zhou2025pay} study the role of small weights and show that they can have a non-trivial impact on model behavior and generalization. 
While their analysis focuses on the magnitude of weights rather than the singular value structure, it complements our perspective that low-energy components of a model or its representations may play a significant role in stability. 
Investigating the relationship between small singular directions and small weights in continual learning settings is an interesting direction for future work.

\section{Methodology}

\subsection{Problem Definition}

Continual Learning (CL) aims to train a model on a sequence of tasks while maintaining strong performance on previously learned tasks.

We consider a CL setting with $T$ \textbf{sequential tasks}
$\{\mathrm{Task}_1, \mathrm{Task}_2, \dots, \mathrm{Task}_T\}$.
At time step $t$, the model observes a dataset
\[
D_t = (X_t, Y_t) = \{(x_{t,i}, y_{t,i})\}_{i=1}^{N_t},
\]
where $X_t$ and $Y_t$ denote the inputs and labels of task $t$, respectively, and $N_t$ is the number of samples in $D_t$. Each input-label pair satisfies $x_{t,i} \in \mathbb{R}^{d}$ and $y_{t,i} \in \mathbb{R}^{d_{\text{out}}}$.

For ease of exposition, we focus on a single linear layer with parameter matrix $\W \in \mathbb{R}^{d \times d_{\text{out}}}$. Let $\W_t$ denote the parameters obtained after training on task $t$. Although we present the formulation for one linear layer, our method applies to an arbitrary number of linear layers in deep networks and can be implemented independently and in parallel across layers, similar in spirit to LoRA-style parameterizations.

Tasks are learned sequentially, and at step $t$ only data from $D_t$ is available for training.

The goal of CL is to learn parameters $\W_T$ that perform well on all tasks $\{1, \dots, T\}$. A common formulation of the objective is
\begin{equation}
    \W = \arg\min_{\W} \sum_{t=1}^{T} \mathcal{L}_t(\W),
\end{equation}
where $\mathcal{L}_t(\W)$ denotes the loss associated with task $t$. In practice, this joint objective cannot be optimized directly due to the sequential data constraint.

The central difficulty in this setting is catastrophic forgetting: updating $\W$ to minimize $\mathcal{L}_t$ may significantly degrade performance on previously learned tasks $\mathcal{L}_1, \dots, \mathcal{L}_{t-1}$.

In this work, we mitigate catastrophic forgetting by constraining parameter updates at each training step, restricting them to subspaces that minimize interference with previously learned representations while still allowing adaptation to new tasks.

\subsection{Our Proposed Method}

Following the problem definition, continual learning can be viewed as balancing stability and plasticity. 
For task $t$, we aim to adapt the model to new data while limiting interference with previously learned tasks.

Assume the network has $L$ layers. Let
\[
\Delta \W_t^{(l)} = \W_t^{(l)} - \W_{t-1}^{(l)}
\]
denote the weight update at layer $l$ when learning task $t$, and let
$
\mathcal{I}_t^{(l)}
$
denote the set of inputs at layer $l$ collected from tasks $1,2,\dots,t-1$.

\paragraph{Stability Constraint (Output Preservation).}

To preserve performance on earlier tasks, we require that learning task $t$ does not significantly alter the outputs induced by previous inputs. This can be expressed as the per-input constraint

\begin{equation}
\label{eq:constraint_output}
\left\|
x^{(l)\top} \Delta \W_t^{(l)}
\right\|_2^2
\le
\varepsilon,
\quad
\forall l \in [L], \;
\forall x^{(l)} \in \mathcal{I}_t^{(l)},
\end{equation}

where $\varepsilon$ controls the allowable output perturbation.

Evaluating this constraint requires access to previous task inputs. In our implementation, such access is limited to a single forward pass per task to construct the subspace used to constrain updates. Exploring fully memory-free alternatives is left for future work.

\paragraph{Plasticity Objective (New Task Learning).}

Under this stability constraint, the learning problem for task $t$ can be formulated conceptually as

\begin{equation}
\label{eq:constrained_objective}
\begin{aligned}
\min_{\W_t} \quad & \mathcal{L}_{CE}(D_t, \W_t) \\
\text{s.t.} \quad
&
\left\|
x^{(l)\top} \Delta \W_t^{(l)}
\right\|_2^2
\le
\varepsilon,
\quad
\forall l \in [L],\;
\forall x^{(l)} \in \mathcal{I}_t^{(l)}.
\end{aligned}
\end{equation}

We emphasize that Eq.~\eqref{eq:constrained_objective} serves as a conceptual formulation of the stability–plasticity trade-off. Rather than solving this constrained problem directly, we introduce a structured parameterization of $\Delta \W_t^{(l)}$ that satisfies the stability constraint by construction.

\paragraph{Construction of the Stability Subspace.}

For clarity, we describe the construction for a single layer and omit the layer index.

Let
\[
I_t = [X_1 : X_2 : \dots : X_{t-1}]
\in
\mathbb{R}^{d \times \sum_{i=1}^{t-1} N_i}
\]
denote the concatenation of all previous task inputs at the considered layer.

We compute its singular value decomposition:

\begin{equation}
\label{eq:svd_Xt}
I_t = \Tilde{U}_t \Sigma_t \Tilde{V}_t^\top,
\end{equation}

where the singular values satisfy
\[
\sigma_{t,1} \ge \sigma_{t,2} \ge \dots \ge \sigma_{t,d} \ge 0.
\]

Directions associated with small singular values correspond to directions with low energy in previously observed inputs. Restricting updates to these directions reduces interference with earlier tasks.

\paragraph{Selecting the Small-Singular-Value Subspace.}

Let $\varepsilon_1 > 0$ be a threshold and define

\begin{equation}
j = \min
\left\{
i :
\sigma_{t,i}
\le
\varepsilon_1 \|I_t\|_F
\right\}.
\end{equation}

The singular vectors
$
\{u_{t,j}, \dots, u_{t,d}\}
$
span an approximate null subspace. Let $U_t =
[u_{t,j} : \dots : u_{t,d}]
$
be the matrix formed by these columns. We parameterize the update as
\begin{equation}
\label{eq:update_param}
\Delta W_t = U_t V_t,
\end{equation}
where $U_t$ is fixed and only $V_t$ is trainable.

\paragraph{Explicit Stability Bound.}

Consider any previous input $x \in \mathcal{I}_t$. 
Since each such $x$ corresponds to a column of $I_t$, there exists a standard basis vector $e_i$ such that

\[
x^\top = e_i^\top I_t^\top.
\]

Using Eq.~\eqref{eq:svd_Xt},
$
I_t^\top
=
\Tilde{V}_t
\Sigma_t
\Tilde{U}_t^\top.
$
Thus,
$
x^\top \Delta W_t
=
e_i^\top
\Tilde{V}_t
\Sigma_t
\Tilde{U}_t^\top
U_t
V_t.
$
Because $U_t$ contains only singular vectors corresponding to indices $j,\dots,d$,
$
\Tilde{U}_t^\top U_t
=
\begin{bmatrix}
0 \\
I
\end{bmatrix},
$
so only the small singular values remain active. Denoting the diagonal matrix of these values by $\Sigma_{\text{small}}$, we obtain
$
x^\top \Delta W_t
=
e_i^\top
\Tilde{V}_t
\Sigma_{\text{small}}
V_t.
$
Taking norms,
$
\|x^\top \Delta W_t\|_2
\le
\|\Sigma_{\text{small}}\|_2
\|V_t\|_2
$
(notice that $\Tilde{V}_t$ is orthogonal), with 
$
\|\Sigma_{\text{small}}\|_2
\le
\varepsilon_1 \|I_t\|_F
$ 
(by construction). Therefore, for every previous input $x$,
\begin{equation}
\|x^\top \Delta W_t\|_2
\le
\varepsilon_1 \|I_t\|_F \|V_t\|_2.
\end{equation}
If we enforce
\begin{equation}
\label{eq:V_bound}
\|V_t\|_2
\le
\frac{\sqrt{\varepsilon}}
{\varepsilon_1 \|I_t\|_F},
\end{equation}
then
$
\|x^\top \Delta W_t\|_2^2
\le
\varepsilon,
\quad
\forall x \in \mathcal{I}_t.
$
Hence, the stability constraint in Eq.~\eqref{eq:constraint_output}
is satisfied for every previous input.

\paragraph{Practical Enforcement.}
In practice, the bound in Eq.~\eqref{eq:V_bound} can be enforced via standard weight decay on $V_t$ during training. 
Thus, by construction, updates lie in an approximate null subspace of previous inputs, while their magnitudes remain controlled, ensuring bounded interference across tasks.
The complete NESS training procedure is summarized in Algorithm~\ref{algo:NESS_method}.

\begin{algorithm}[ht]
    \caption{NESS}
    \label{algo:NESS_method}
    \begin{normalsize}
        \DontPrintSemicolon
        \SetKwInOut{KwIn}{Input}
        \SetKwInOut{KwOut}{Output}
        \SetKwFunction{FGetUV}{\texttt{GetUV}}
        \SetKwProg{Pn}{function}{}{}

        \KwIn{$X = [x_i] \in \mathbb{R}^{d \times N}$ (input matrix with $N$ samples), $\varepsilon_1$ (threshold), $d_{\text{out}}$ (output dimension)}
        \KwOut{$U_t$ (frozen orthogonal bases), $V_t$ (trainable matrix)}

        \Pn{\FGetUV{$X, \varepsilon_1, d_{\text{out}}$}}{
            $C \leftarrow XX^\top$ \tcp*{covariance matrix}
            $\mathbf{U}, \mathbf{\Lambda} \leftarrow \text{eig}(C)$ \\
            \tcp{Eigenvalues are sorted in descending order}
            $\{\sigma_1,\dots,\sigma_d\} \leftarrow \sqrt{\text{diag}(\mathbf{\Lambda})}$ \\
            $\{u_1,\dots,u_d\} \leftarrow \text{Column}(\mathbf{U})$ \\
            
            $j = \min\{ k \;:\; \sigma_k \le \varepsilon_1 \|X\|_F \}$ \hfill\textcolor{gray}{\small{// select small-singular-value directions}}\\
            
            $U_t \leftarrow [u_j, \dots, u_d]$ \hfill\textcolor{gray}{// $U_t \in \mathbb{R}^{d \times (d-j+1)}$}\\
            
            $V_t \leftarrow \mathbf{0} \in \mathbb{R}^{(d-j+1) \times d_{\text{out}}}$ \\
            
            \Return $U_t, V_t$
        }
    \end{normalsize}

    \vspace{0.25cm}
    \begin{normalsize}
        \DontPrintSemicolon
        \SetKwInOut{KwIn}{Input}
        \SetKwInOut{KwOut}{Output}
        \SetKwFunction{FTrain}{\texttt{train}}
        \SetKwProg{Pn}{function}{}{}

        \KwIn{$t$ (current task index), $D_t = (X_t, Y_t)$ (task dataset), $\varepsilon_1$ (threshold), \\
        $\mathcal{L} = [1,2,\dots,L]$ (layers), $\mathcal{W} = [W^{(l)}]_{l=1}^L$ (current weights)}
        
        \KwOut{Updated weights $\mathcal{W}$}

        \Pn{\FTrain{$t, D_t, \varepsilon_1, \mathcal{L}, \mathcal{W}$}}{
            \For{each layer $l \in \mathcal{L}$}{
                $I^{(l)} \leftarrow$ forward pass to collect inputs of layer $l$ on $D_t$ \\
                $(U^{(l)}, V^{(l)}) \leftarrow \texttt{GetUV}(I^{(l)}, \varepsilon_1, \text{output\_dim of layer } l)$
            }

            \tcp{Train only $V^{(l)}$ (backbone frozen); enforce $\|V^{(l)}\|$ bound via regularization}
            $\{V^{(l)}\}_{l=1}^L \leftarrow 
            \arg\min_{\{V\}} 
            \mathcal{L}_{CE}\!\left(D_t,\; \{W^{(l)} + U^{(l)}V^{(l)}\}_{l=1}^L\right)$
            
            \For{each layer $l \in \mathcal{L}$}{
                $W^{(l)} \leftarrow W^{(l)} + U^{(l)} V^{(l)}$ \hfill\textcolor{gray}{// merge adapter}
            }

            \Return $\mathcal{W}$
        }
    \end{normalsize}

    \vspace{0.25cm}
    \begin{normalsize}
        \DontPrintSemicolon
        \SetKwInOut{KwIn}{Input}
        \SetKwInOut{KwOut}{Output}
        \SetKwFunction{FFull}{\texttt{full\_training}}
        \SetKwProg{Pn}{function}{}{}

        \KwIn{$T$ (number of tasks), $D = \{D_t\}_{t=1}^T$, $\varepsilon_1$, $\mathcal{L}$ (layers)}
        \KwOut{Final weights $\mathcal{W}$}

        \Pn{\FFull{$T, D, \varepsilon_1, \mathcal{L}$}}{
            Initialize $\mathcal{W} = [W^{(l)}]_{l=1}^L$ \hfill\textcolor{gray}{// pretrained or zero initialization}

            \For{$t = 1$ to $T$}{
                $\mathcal{W} \leftarrow \texttt{train}(t, D_t, \varepsilon_1, \mathcal{L}, \mathcal{W})$
            }

            \Return $\mathcal{W}$
        }
    \end{normalsize}
\end{algorithm}

\subsection{Training Paradigm}

In this subsection, we describe the training procedure of NESS. 
Assume that the network contains $L$ linear or convolutional layers.

\texttt{GetUV} \textbf{function (lines 1–10 of Algorithm~\ref{algo:NESS_method}).}  
This function constructs the frozen basis $U_t$ and the trainable matrix $V_t$ used to parameterize layer updates.

Given an input matrix $X$, we compute the eigen-decomposition of the outer product
\[
XX^\top = \sum_i x_i x_i^\top,
\]
which is equivalent to computing the left singular vectors $\mathbf{U}$ of $X$. The eigenvalues (corresponding to squared singular values) are sorted in descending order. We then identify index $j$ as the smallest index such that
$
\sigma_j \le \varepsilon_1 \|X\|_F.
$
The matrix $U_t$ is constructed from the columns of $\mathbf{U}$ corresponding to indices $j,\dots,d$, i.e., the directions associated with small singular values. These directions correspond to low-energy components of previously observed inputs and form an approximate null subspace. This procedure is inspired by efficient low-rank compression initialization~\citep{zaccone2026flexranknestedlowrankknowledge}.

Intuitively, large singular values correspond to directions with high variance in past data, and updates along these directions are more likely to interfere with previously learned representations. Restricting updates to the subspace associated with small singular values therefore limits interference while preserving sufficient flexibility for adaptation.

Finally, we initialize a trainable matrix $V_t$ such that the layer update is parameterized as
$
\Delta W_t = U_t V_t.
$
The \texttt{GetUV} function returns the frozen basis $U_t$ and the trainable matrix $V_t$.
For computational efficiency, we compute $XX^\top$ in an online fashion by accumulating rank-one updates $\sum_i x_i x_i^\top$, without storing the full input matrix $X$. This avoids explicitly computing and storing right singular vectors and reduces memory usage, as only the covariance matrix and its eigen-decomposition are required.

The full training procedure (function \texttt{full\_training}, lines 20-24 of Algorithm~\ref{algo:NESS_method}) consists of two phases: training the first task $1$, and training tasks $2$ through $T$.

\textbf{Training task $1$.}  
The first task is trained without subspace constraints. We perform standard gradient descent on all layers using dataset $D_1$. After training, we obtain the backbone weights
\[
\mathcal{W} = [\W_1^{(1)}, \W_1^{(2)}, \dots, \W_1^{(L)}].
\]

\textbf{Training tasks $2$ to $T$.}  
For each subsequent task $t$, we adapt the linear and convolutional layers using the structured parameterization described above (function \texttt{train}, lines 11-19 of Algorithm~\ref{algo:NESS_method}).

For each layer $l$, we perform a forward pass to collect its input matrix $I^{(l)}$. Using \texttt{GetUV}, we construct $U_t^{(l)}$ and initialize $V_t^{(l)}$. Only $V_t^{(l)}$ is trainable, while $U_t^{(l)}$ remains fixed. The update for layer $l$ is parameterized as
$
\Delta \W_t^{(l)} = U_t^{(l)} V_t^{(l)}.
$

All matrices $\{V_t^{(l)}\}_{l=1}^L$ are trained jointly using the current task dataset $D_t$ and the cross-entropy loss. During optimization, norm regularization (e.g., weight decay) is applied to $V_t^{(l)}$ to enforce the stability bound derived in the previous subsection. By construction, this guarantees that the induced output perturbation on previous inputs remains bounded.

After optimization, the updates are merged into the backbone weights $\W_t^{(l)} \leftarrow \W_{t-1}^{(l)} + U_t^{(l)} V_t^{(l)}.$
Finally, since our experiments focus on image classification, a task-specific classification head is used. This head is trained without the subspace constraint to allow full adaptation to the new task.

\section{Experiments}
\subsection{Experimental Setup}

\begin{table}[ht!]
\centering
\scriptsize
\definecolor{lightblue}{RGB}{173, 216, 230}
\definecolor{yellow}{RGB}{253, 218, 13}
\caption{The averaged accuracy (ACC) and backward transfer (BWT) over all tasks on different datasets (with the best learning rate scenario). Accuracy and BWT (mean $\pm$ std) are reported based on 5 different seeds. The best BWTs are in bold, and the second-best BWTs are underlined, and all the BWTs above $-1\%$ have a \colorbox{yellow}{yellow background}. For notations: $^{\dagger}$ denotes that the baseline has no previous code/result for all datasets, and $^{\ddagger}$ denotes the results taken from \cite{Apolinario_2025_ICCV}. For a more detailed result, please see Appendix \ref{apd:addition_results}.}
\vspace{0.5cm}
\label{tab:main_results}
\begin{tabular}{lcccccc}
\toprule
\multirow{2}{*}{\textbf{Method}} &
\multicolumn{2}{c}{CIFAR-100 (10 Tasks)} &
\multicolumn{2}{c}{5-datasets (5 Tasks)} &
\multicolumn{2}{c}{MiniImageNet (20 Tasks)}  \\ 
\cmidrule(lr){2-3} \cmidrule(lr){4-5} \cmidrule(lr){6-7}
 & ACC(\%) & \textbf{BWT(\%)$\uparrow$} & ACC(\%) & \textbf{BWT(\%)$\uparrow$} & ACC(\%) & \textbf{BWT(\%)$\uparrow$} \\
\midrule
OWM$^{\ddagger}$~\cite{zeng2019continual}     & 50.94 $\pm$ 0.60 & -30 $\pm$ 1
           & -- & --
           & -- & -- \\
EWC$^{\ddagger}$~\cite{kirkpatrick2017overcoming}     & 68.88 $\pm$ 0.80 & -2 $\pm$ 1
           & 88.64 $\pm$ 0.26 & -4 $\pm$ 1
           & 52.01 $\pm$ 2.53 & -12 $\pm$ 3 \\
HAT$^{\ddagger}$~\cite{serra2018overcoming}     & 72.06 $\pm$ 0.50 & 0 $\pm$ 0
           & 91.32 $\pm$ 0.18 & -1 $\pm$ 0
           & 59.78 $\pm$ 0.57 & -3 $\pm$ 0 \\
A-GEM$^{\ddagger}$~\cite{chaudhryefficient}     & 63.98 $\pm$ 1.22 & -15 $\pm$ 2
           & 84.04 $\pm$ 0.33 & -12 $\pm$ 1
           & 57.24 $\pm$ 0.72 & -12 $\pm$ 1 \\
GPM~\cite{saha2021gradient}     & 71.63 $\pm$ 0.67 & \cellcolor{yellow}-0.28 $\pm$ 0.54
           & 90.61 $\pm$ 0.57 & -1.17 $\pm$ 0.26
           & 63.56 $\pm$ 2.42 & -1.39 $\pm$ 1.37 \\
           
SGP~\cite{saha2023continual}     & 75.99 $\pm$ 0.16 & -1.18 $\pm$ 0.37
           & 90.48 $\pm$ 0.48 & -1.82 $\pm$ 0.13
           & 64.45 $\pm$ 2.18 & \cellcolor{yellow}-0.53 $\pm$ 0.68 \\
TRGP~\cite{lin2022trgp}       & 75.21 $\pm$ 0.32 &  \cellcolor{yellow}\textbf{0.06 $\pm$ 0.17}
           & 92.78 $\pm$ 0.65 &  \cellcolor{yellow}\textbf{-0.09 $\pm$ 0.08}
           & 62.74 $\pm$ 2.13 &  -1.23 $\pm$ 0.77 \\
FS-DGPM$^{\dagger}$~\cite{deng2021flattening}       & 74.10 $\pm$ 0.09 & -3.03 $\pm$ 0.31
           & -- & --
           & -- & -- \\        
DFGP~\cite{Yang_2023_ICCV, yang_2025_tpami} (mixup=0.01)     & 73.24 $\pm$ 0.24 & \cellcolor{yellow}-0.95 $\pm$ 0.18
           & 91.47 $\pm$ 0.22 & -2.27 $\pm$ 0.31
           & 68.50 $\pm$ 1.50 & \cellcolor{yellow}\underline{-0.11  $\pm$ 1.36} \\
DFGP~\cite{Yang_2023_ICCV, yang_2025_tpami} (mixup=0.05)     & 73.77 $\pm$ 0.33 & -1.03 $\pm$ 0.41
           & 90.22 $\pm$ 0.46 & -3.87 $\pm$ 0.53
           & 68.64 $\pm$ 2.25 & \cellcolor{yellow}-0.43 $\pm$ 1.60 \\
DFGP~\cite{Yang_2023_ICCV, yang_2025_tpami} (mixup=0.001)     & 73.32 $\pm$ 0.50 & -1.09 $\pm$ 0.49
           & 91.59 $\pm$ 0.50 & -1.86 $\pm$ 0.25
           & 67.75 $\pm$ 1.81 & -1.11 $\pm$ 1.50 \\

DFGP~\cite{Yang_2023_ICCV, yang_2025_tpami} (mixup=0.0001)     & 73.16 $\pm$ 0.46 & -1.11 $\pm$ 0.52
           & 91.51 $\pm$ 0.15 & -1.70 $\pm$ 0.42 
           & 68.39 $\pm$ 1.07 & \cellcolor{yellow}-0.16 $\pm$ 1.42 \\

\midrule
\textbf{NESS (with SAM)}     & 72.56 $\pm$ 0.07 & \cellcolor{yellow}-0.17 $\pm$ 0.51
           & 90.98 $\pm$ 0.07 & \cellcolor{yellow}-0.86 $\pm$ 0.28
           & 63.48 $\pm$ 1.38 & \cellcolor{yellow}-0.26 $\pm$ 0.67 \\

\textbf{NESS (with SGDm: m=0.9) }    & 72.46 $\pm$ 0.26 & \cellcolor{yellow}\underline{0.03 $\pm$ 0.40}
           & 90.20 $\pm$ 0.47 & \cellcolor{yellow}\underline{-0.58 $\pm$ 0.15}
           & 63.72 $\pm$ 0.46 & \cellcolor{yellow}\textbf{0.41 $\pm$ 0.58} \\
\bottomrule
\end{tabular}
\end{table}

\textbf{Datasets. } Following previous works \citep{lin2022trgp, saha2023continual, Yang_2023_ICCV, Apolinario_2025_ICCV}, we evaluated NESS on 3 different image classification benchmark datasets: \textbf{CIFAR-100} \citep{krizhevsky2009learning} (split to 10 tasks), \textbf{5-datasets} \citep{vinyals2016matching} (5 tasks, with each task being 1 task), and \textbf{MiniImageNet} \citep{ebrahimi2020adversarial} (split to 20 tasks). The datasets' statistics are described in Appendix~\ref{apd:exp_settings}. For network architecture, we use networks similar to previous works, including a 5-layer AlexNet for CIFAR-100 and a Reduced ResNet18 for 5-datasets and MiniImageNet.

\textbf{Baselines. } We compared NESS with 4 traditional baselines: OWM \citep{zeng2019continual}, EWC \citep{kirkpatrick2017overcoming}, HAT \citep{serra2018overcoming}, and A-GEM \citep{chaudhryefficient}; and 5 strong orthogonal-based baselines: GPM \citep{saha2021gradient}, SGP \citep{saha2023continual}, TRGP \citep{lin2022trgp}, FS-DGPM \citep{deng2021flattening}, DFGP \citep{Yang_2023_ICCV, yang_2025_tpami}. Specifically for DFGP, we conducted experiments using their 3 or 4 mixup weight rates, depending on the dataset and its original settings. 

\textbf{Metrics.} Following previous work of \citet{saha2021gradient}, we evaluated the continual training performance using 2 metrics to evaluate a CL model: Average Accuracy and Backward Transfer, \textbf{in which we focus mainly on Backward Transfer.}
\begin{itemize}
    \item \textbf{Average Accuracy (ACC)}: represents the average test accuracy of the model trained on all tasks (defined as $\text{ACC} = \frac{1}{T}\sum_{i=1}^{T}A_{T, i}$).
    \item \textbf{Backward Transfer (BWT)}: measures the forgetting of old tasks (defined as $\text{BWT} = \frac{1}{T-1}\sum_{i=1}^{T-1}(A_{T, i} - A_{i, i})$).
\end{itemize}
with $T$ denotes number of tasks and $A_{t, i}$ denotes the accuracy tested on task $i$ after training with task $t$. 

\textbf{Settings for NESS. }
For NESS, we conducted experiments using three optimizers: SAM \citep{foretsharpness} (used in DFGP) and SGD with momentum (SGDm, momentum rate = 0.9).

\subsection{Experimental Results}

\textbf{Performance.} 
The summary of our experiment results is shown in Table~\ref{tab:main_results} (for a more detailed version, please see Appendix~\ref{apd:addition_results}). 
In addition, Figure~\ref {fig:mini_seed37} (and additional figures in Appendix~\ref{apd:addition_results}) demonstrate the forgetting rate across all tasks for NESS and other baselines (fewer red cells indicate better performance). As shown in the results of Table~\ref{tab:main_results} and the above two figures, we can see that NESS is efficient and stable for continual learning:
\begin{itemize}
    \item \textbf{Efficiency:} The results for NESS using SGD with momentum achieve the best BWT rate on three different datasets, which is comparable to the best baseline (TRGP) on CIFAR-100 \& 5-datasets, and also beat the best baseline (DFGP) on MiniImageNet.
    \item \textbf{Stability:} The performance of NESS with different optimizers are similar and can be compared with the baselines. Moreover, our method always achieves a BWT rate greater than -1\% or even with positive BWT (see the \textit{yellow cells} in Table~\ref{tab:main_results}), which is better than all baselines where at least one of the settings produces a BWT rate less than -1\%, or even -3\% and significantly hurts the CL model.
\end{itemize}

\textbf{Efficiency Analysis.} 
NESS is efficient by design. Indeed, because the trainable matrix is always guaranteed to be smaller than or equal to the size of the corresponding layer (with the construction of a dynamic trainable matrix in every linear/convolution layer with threshold $\varepsilon_1$), NESS guarantees that the total number of trainable parameters will be much smaller than the total network parameters.
It is also easy to see that the threshold magnitude and the number of trainable parameters are proportional. The lower the threshold $\varepsilon_1$, the more efficient the training parameters we achieved.

\section{Conclusions}

In this work, we revisit the use of small singular values in orthogonal-based continual learning methods. Then, we propose NESS, a new CL algorithm that uses orthogonal bases corresponding to the small singular values of previous input representations. By constructing a trainable matrix lying in the null space of the input data and training on the current task data, NESS can prevent catastrophic forgetting while achieving strong backward transfer across tasks. Subsequently, we analyze the theoretical guarantees and the efficiency of our proposed model. Finally, extensive experimental results across three datasets demonstrate the promising performance of NESS, with stable performance and low or even negative forgetting rates, highlighting the potential impact of small singular values in continual learning. We also leave open problems related to effectiveness and threshold tuning for future studies.

\bibliography{references}


\newpage
\appendix
\section{Appendix}

\subsection{Detailed Experiments Settings} \label{apd:exp_settings}

The statistics for three datasets used for continual training in this paper are summarized in Table~\ref{tab:split_stats} and Table~\ref{tab:5datasets_stats}.
\begin{table}[ht!]
\centering
\caption{Split CIFAR-100 and split MiniImageNet datasets statistics.}
\vspace{0.3cm}
\begin{tabular}{l|cc}
\hline
\textbf{Dataset} & \textbf{CIFAR-100} & \textbf{MiniImageNet} \\ \hline
Number of tasks ($T$)              & 10           & 20           \\
Number of classes per task         & 10           & 5            \\
Training samples per task          & 4750         & 2375         \\
Validation samples per task        & 250          & 125          \\
Test samples per task              & 1000         & 500          \\ \hline
\end{tabular}
\label{tab:split_stats}
\end{table}

\begin{table}[ht!]
\centering
\caption{5-Datasets statistics (notice that each dataset is one task).}
\vspace{0.3cm}
\begin{tabular}{l|ccccc}
\hline
\textbf{Dataset} & \textbf{CIFAR10} & \textbf{MNIST} & \textbf{SVHN} & \textbf{Fashion MNIST} & \textbf{notMNIST} \\ \hline
Number of classes      & 10    & 10    & 10    & 10    & 10    \\
Training samples       & 47500 & 57000 & 69595 & 57000 & 16011 \\
Validation samples     & 2500  & 3000  & 3662  & 3000  & 842   \\
Test samples           & 10000 & 10000 & 26032 & 10000 & 1873  \\ \hline
\end{tabular}
\label{tab:5datasets_stats}
\end{table}

The hyperparameters used in our experiments are detailed in Table~\ref{tab:hyperparams}. For more details on the learning rate (lr) and $\varepsilon_1$ in each experiment, please see the Appendix~\ref{apd:addition_results}.
\begin{table}[ht!]
\centering
\caption{List of (general) hyperparameters used in our experiments.}
\vspace{0.3cm}
\begin{tabular}{l|ccc}
\hline
\textbf{Dataset} & \textbf{CIFAR-100} & \textbf{5-Datasets} & \textbf{MiniImageNet} \\ \hline
Batch size ($b$)                       & 64    & 64   & 64   \\
Learning rate decay factor            & 1/2 & 1/3 & 1/2 \\
Patience                               & 6     & 5    & 6    \\
Number of epochs                & 200   & 100  & 100  \\
Momentum (for SGD with momentum) & 0.9   & 0.9  & 0.9  \\
\hline
\end{tabular}
\label{tab:hyperparams}
\end{table}

\subsection{Additional Results} \label{apd:addition_results}

\textbf{The Value of Threshold $\varepsilon_1$. }
For our experiments, we have used different values for the threshold $\varepsilon_1$, ranging from 0.01 to 0.0001. Through our experiments on three different datasets, we observed that for every optimizer, $\varepsilon_1 = 0.001$ or $\varepsilon_1 = 0.0005$ achieved better results.

\textbf{Additional experiments. }
We also conducted an additional experiment on DFGP without using mixup and data augmentation as another baseline for NESS. The results show that without mixup and augmentation technique, the performance of DFGP can still compare with the original DFGP. However, it still can not achieve the BWT rate of NESS.

\textbf{Detail results. }
The detailed results of every experiment are shown as 2 types: (i) \textbf{detailed results} (CIFAR-100 in Table~\ref{tab:seeds_CIFAR100}, 5-datasets in Table~\ref{tab:seeds_FIVE}, MiniImageNet in Table~\ref{tab:seeds_MINI}); and (ii) \textbf{average performance} (CIFAR-100 in Table~\ref{tab:seeds_avg_CIFAR100}, 5-datasets in Table~\ref{tab:seeds_avg_FIVE}, MiniImageNet in Table~\ref{tab:seeds_avg_MINI}). For each experiment, we chose 5 different seeds: 1 (following previous works on CIFAR-100), 2, 3, 4, and 37 (following previous works on 5-datasets and MiniImageNet). 

\begin{figure}[t]
    \centering
    \begin{subfigure}[b]{0.48\textwidth}
        \centering
        \includegraphics[width=\textwidth]{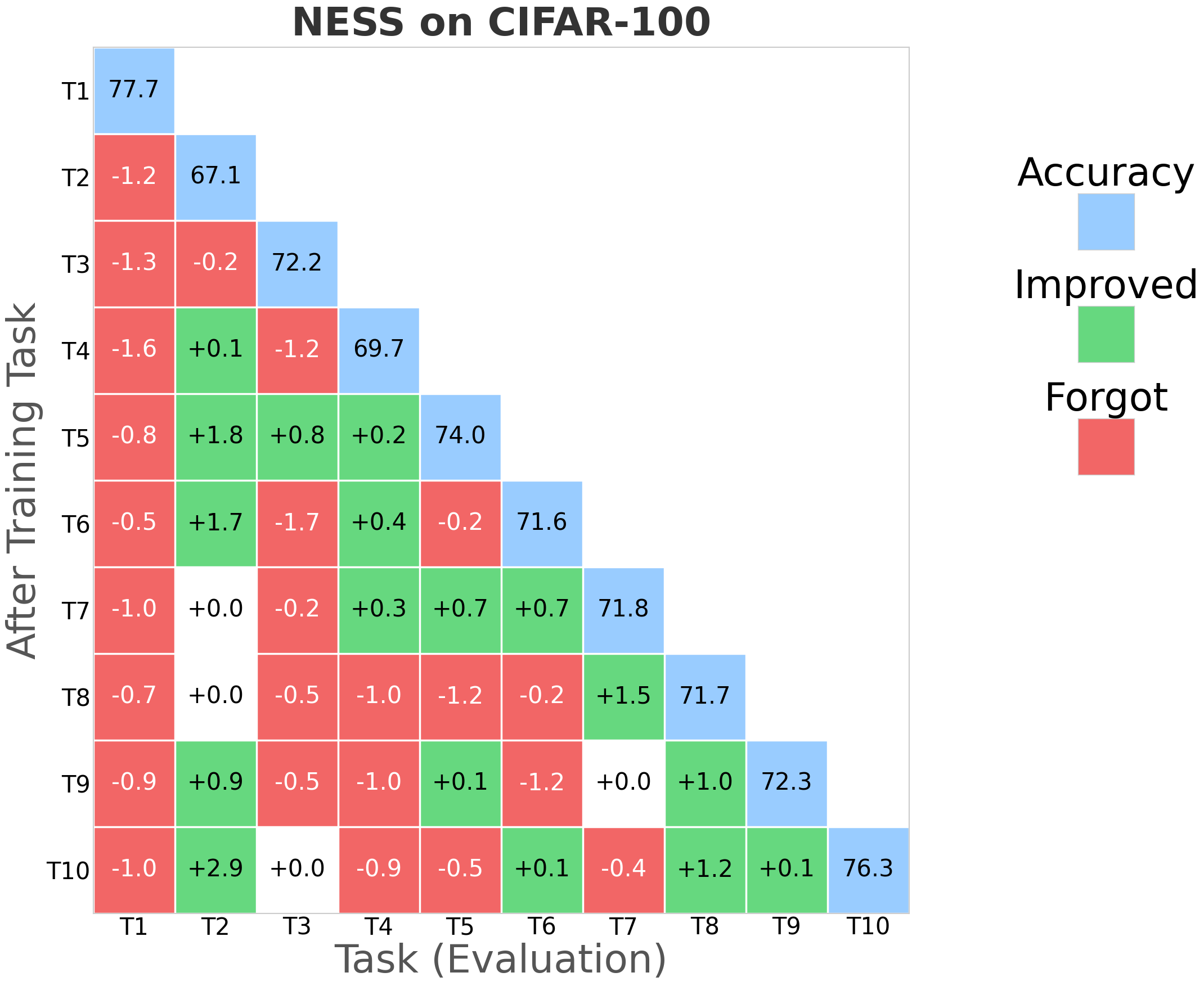}
        \caption{NESS}
    \end{subfigure}
    \hfill
    \begin{subfigure}[b]{0.48\textwidth}
        \centering
        \includegraphics[width=\textwidth]{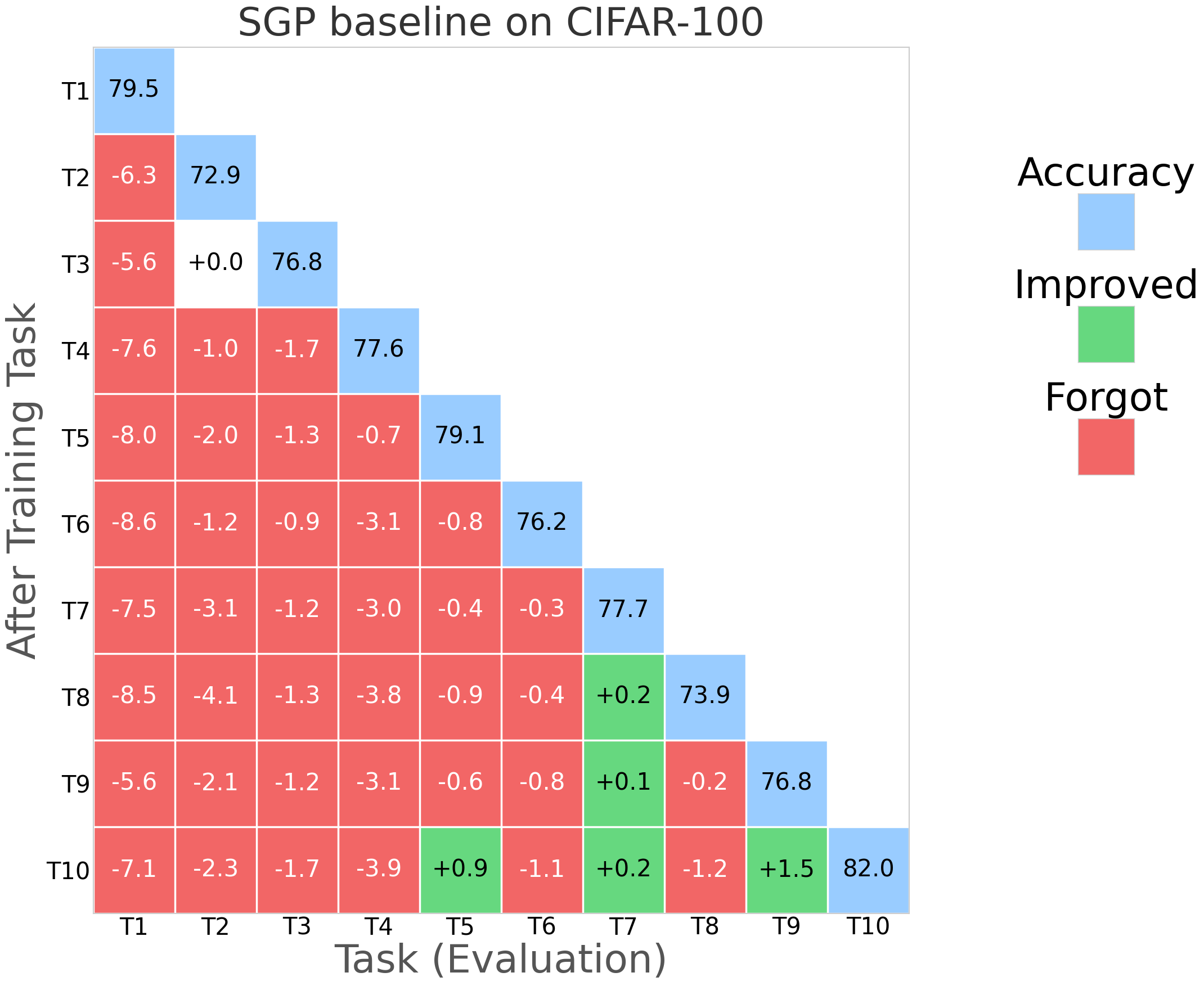}
        \caption{SGP baseline}
    \end{subfigure}
    
    \vspace{0.5cm}
    
    \begin{subfigure}[b]{0.48\textwidth}
        \centering
        \includegraphics[width=\textwidth]{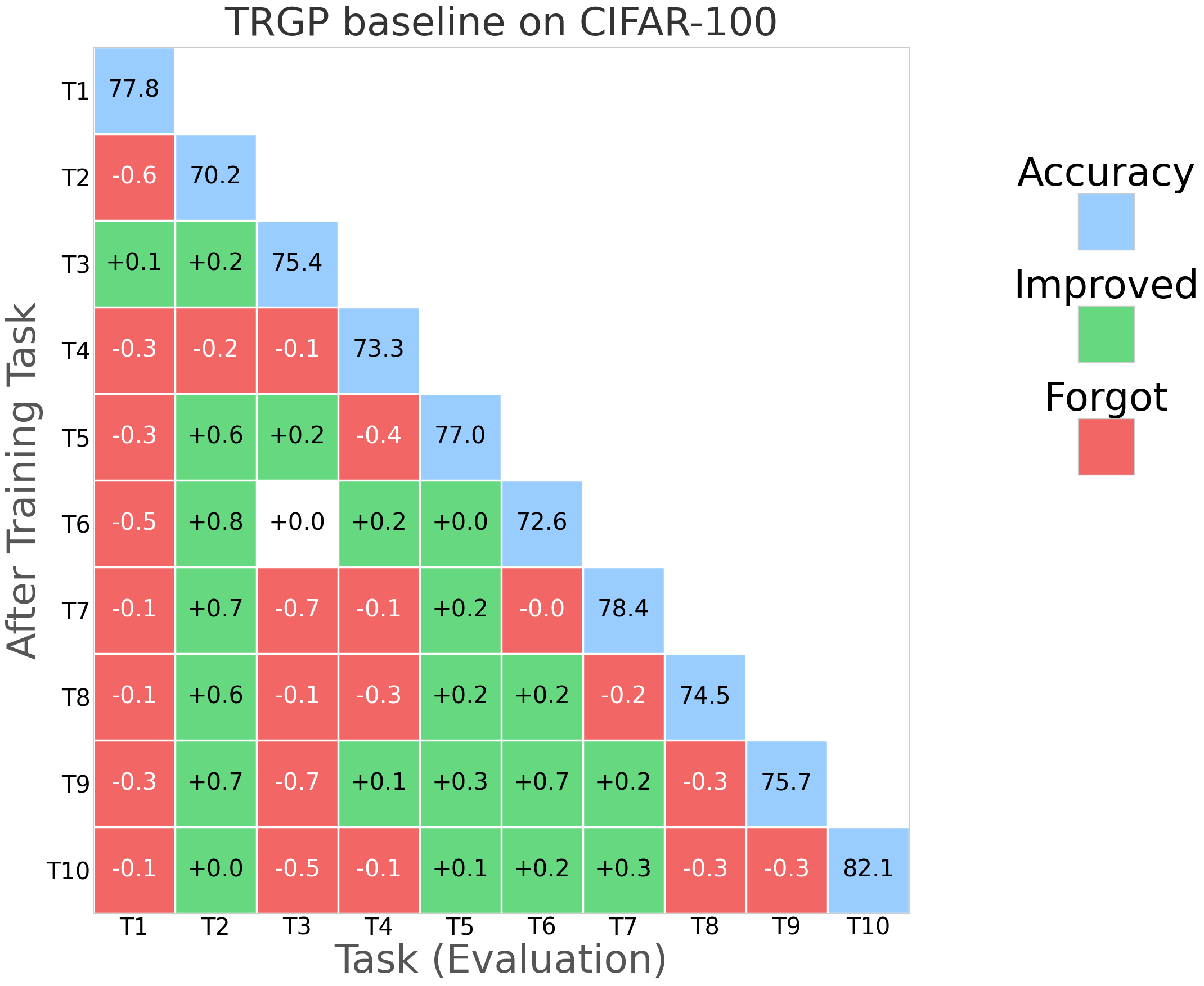}
        \caption{TRGP baseline}
    \end{subfigure}
    \hfill
    \begin{subfigure}[b]{0.48\textwidth}
        \centering
        \includegraphics[width=\textwidth]{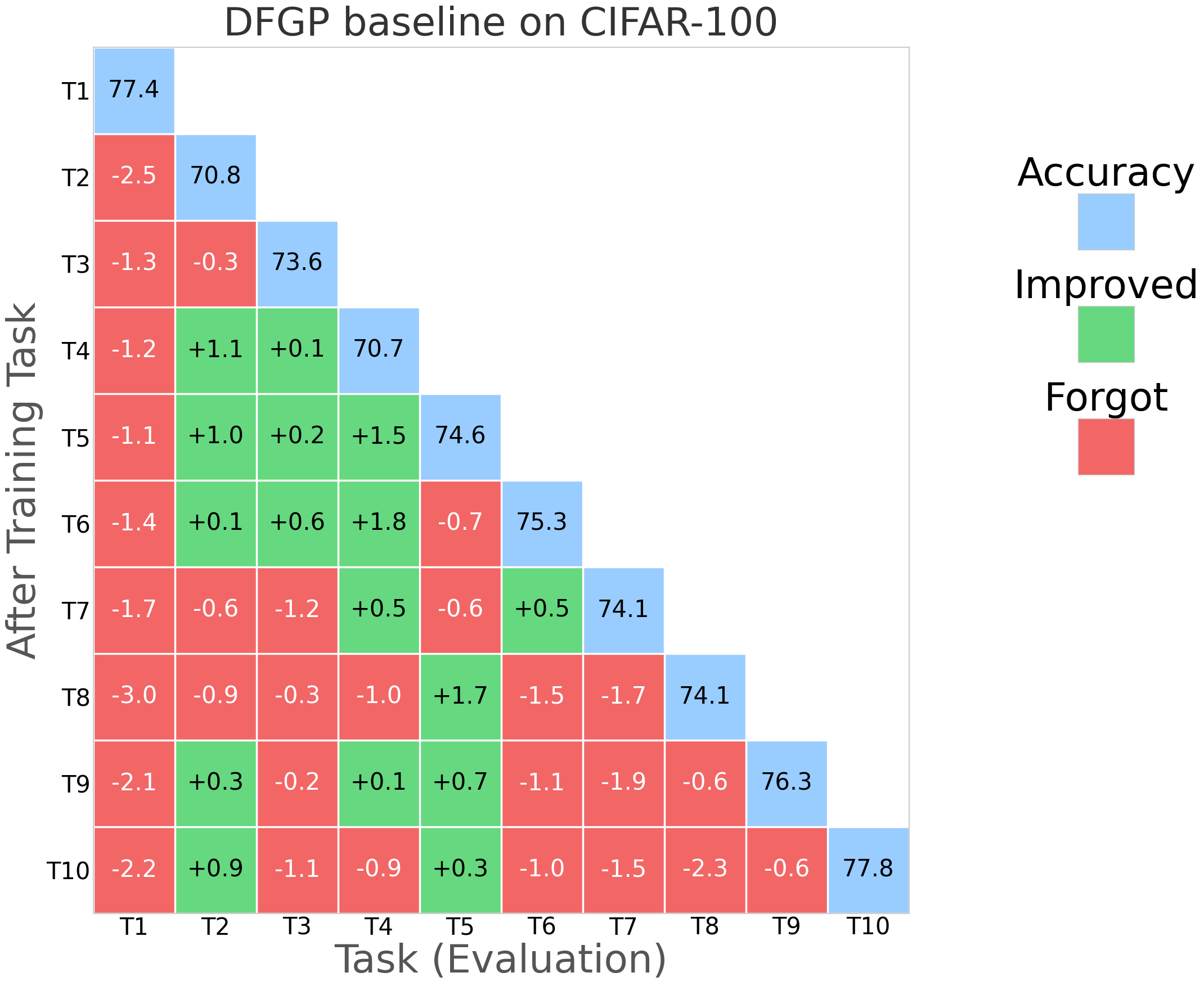}
        \caption{DFGP baseline}
    \end{subfigure}
    \caption{Performance (best setting of each method, focusing on forgetting rate, with green color denotes improving performance and red color denotes losing performance, compared to the time the task firstly trained) of (a) NESS compared to (b) SGP; (c) TRGP; (d) DFGP on CIFAR-100 experiment (seed 2).}
    \label{fig:cifar_seed2}
\end{figure}

\begin{figure}[t]
    \centering
    \begin{subfigure}[b]{0.48\textwidth}
        \centering
        \includegraphics[width=\textwidth]{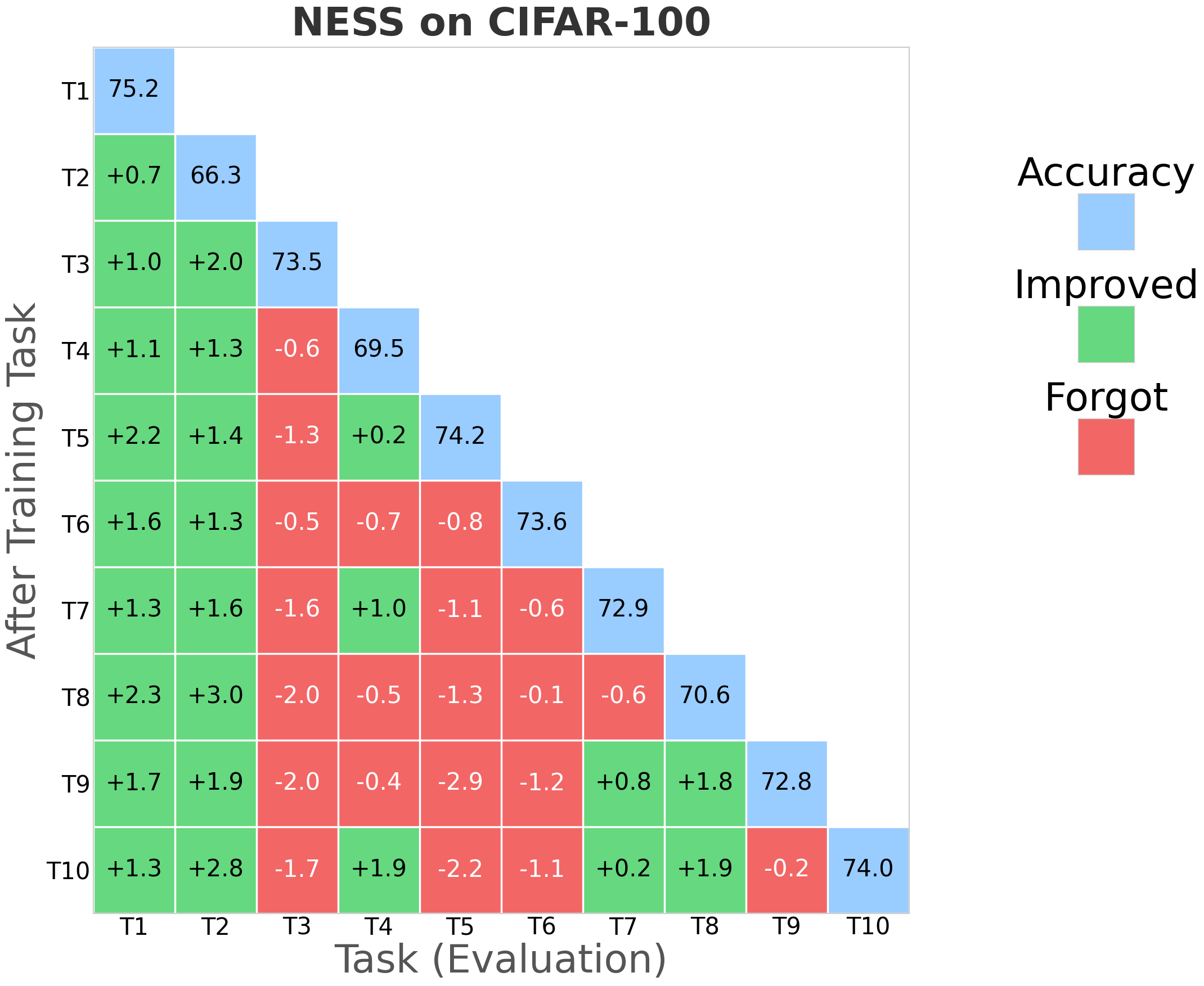}
        \caption{NESS}
    \end{subfigure}
    \hfill
    \begin{subfigure}[b]{0.48\textwidth}
        \centering
        \includegraphics[width=\textwidth]{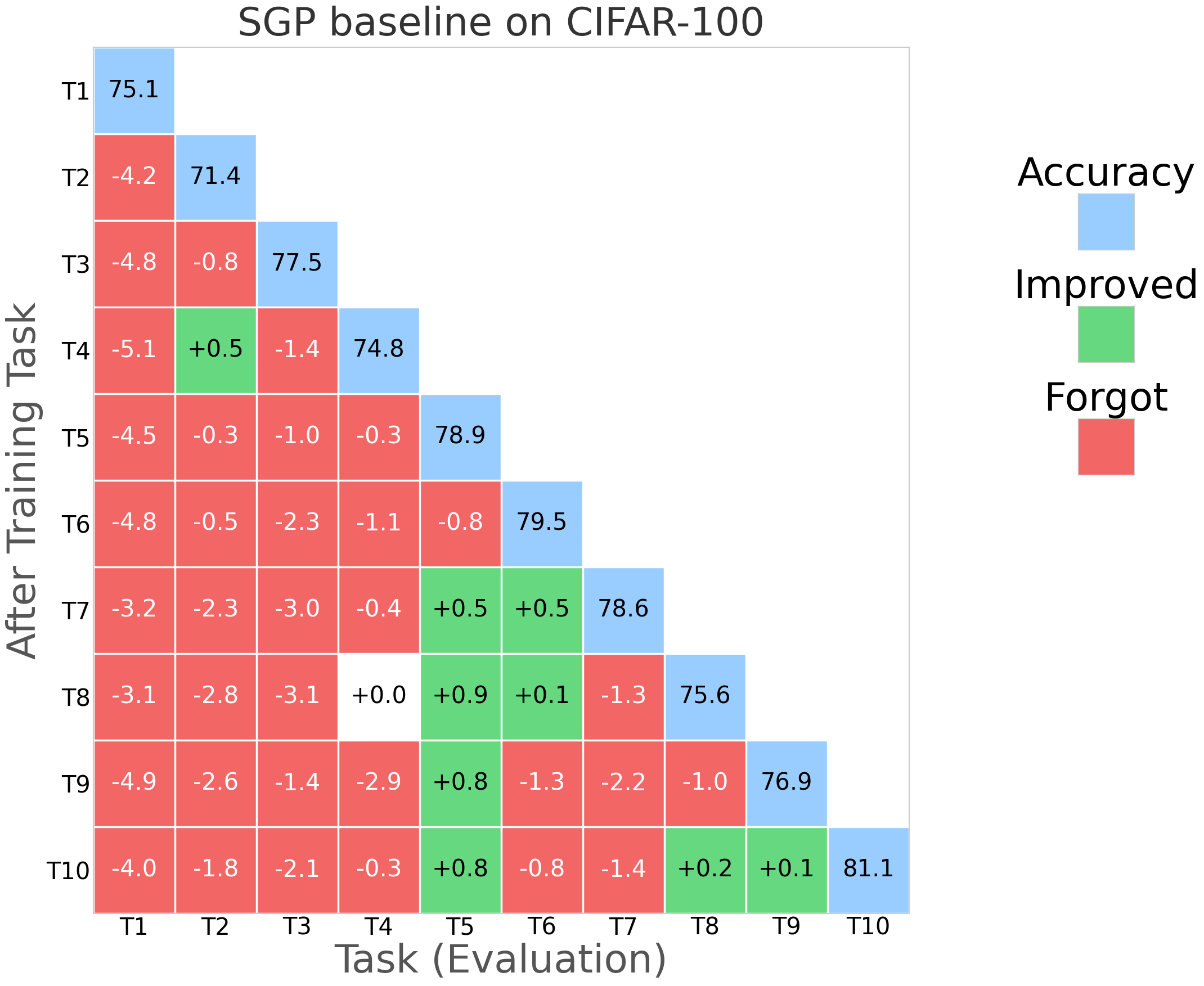}
        \caption{SGP baseline}
    \end{subfigure}
    
    \vspace{0.5cm}
    
    \begin{subfigure}[b]{0.48\textwidth}
        \centering
        \includegraphics[width=\textwidth]{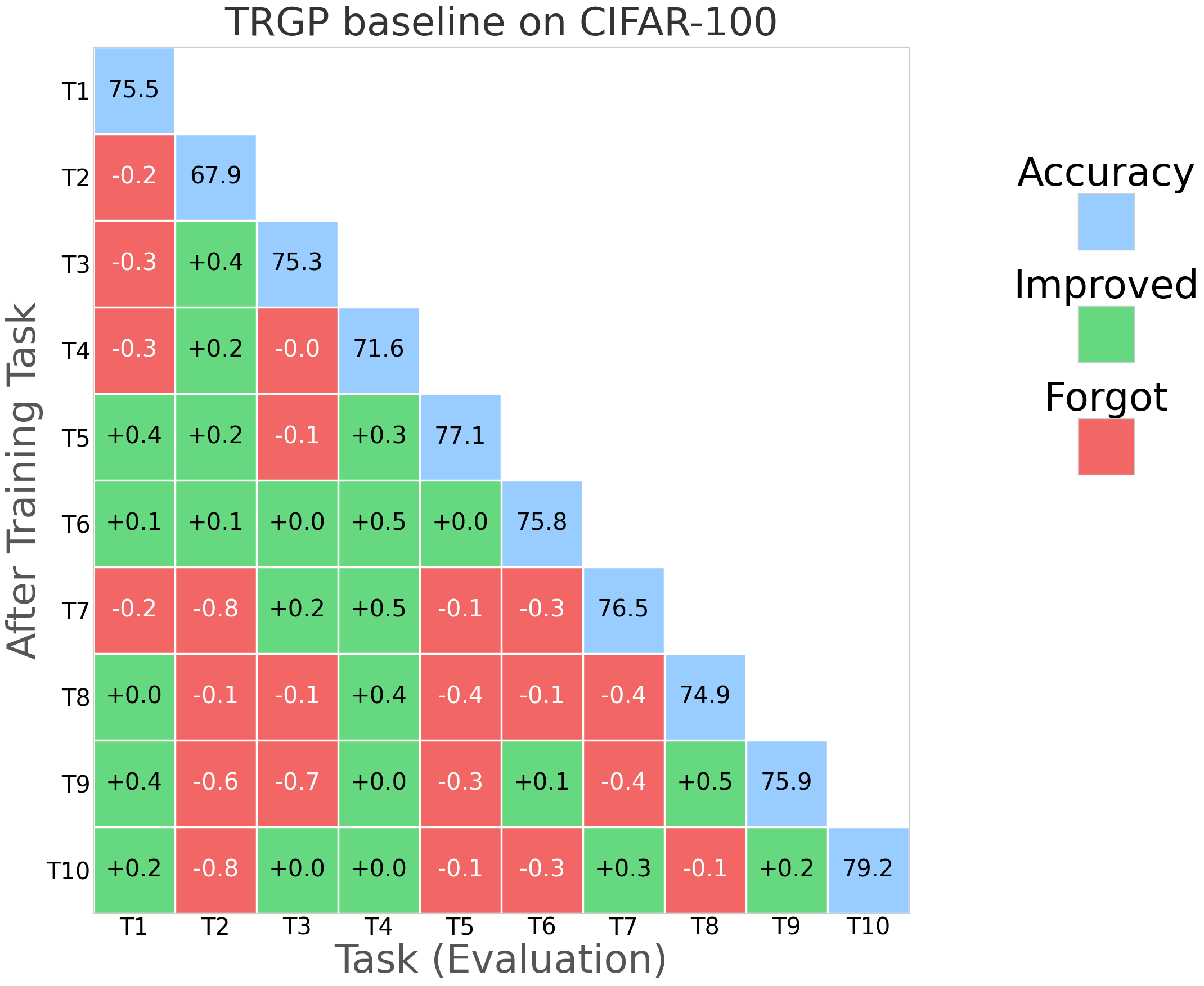}
        \caption{TRGP baseline}
    \end{subfigure}
    \hfill
    \begin{subfigure}[b]{0.48\textwidth}
        \centering
        \includegraphics[width=\textwidth]{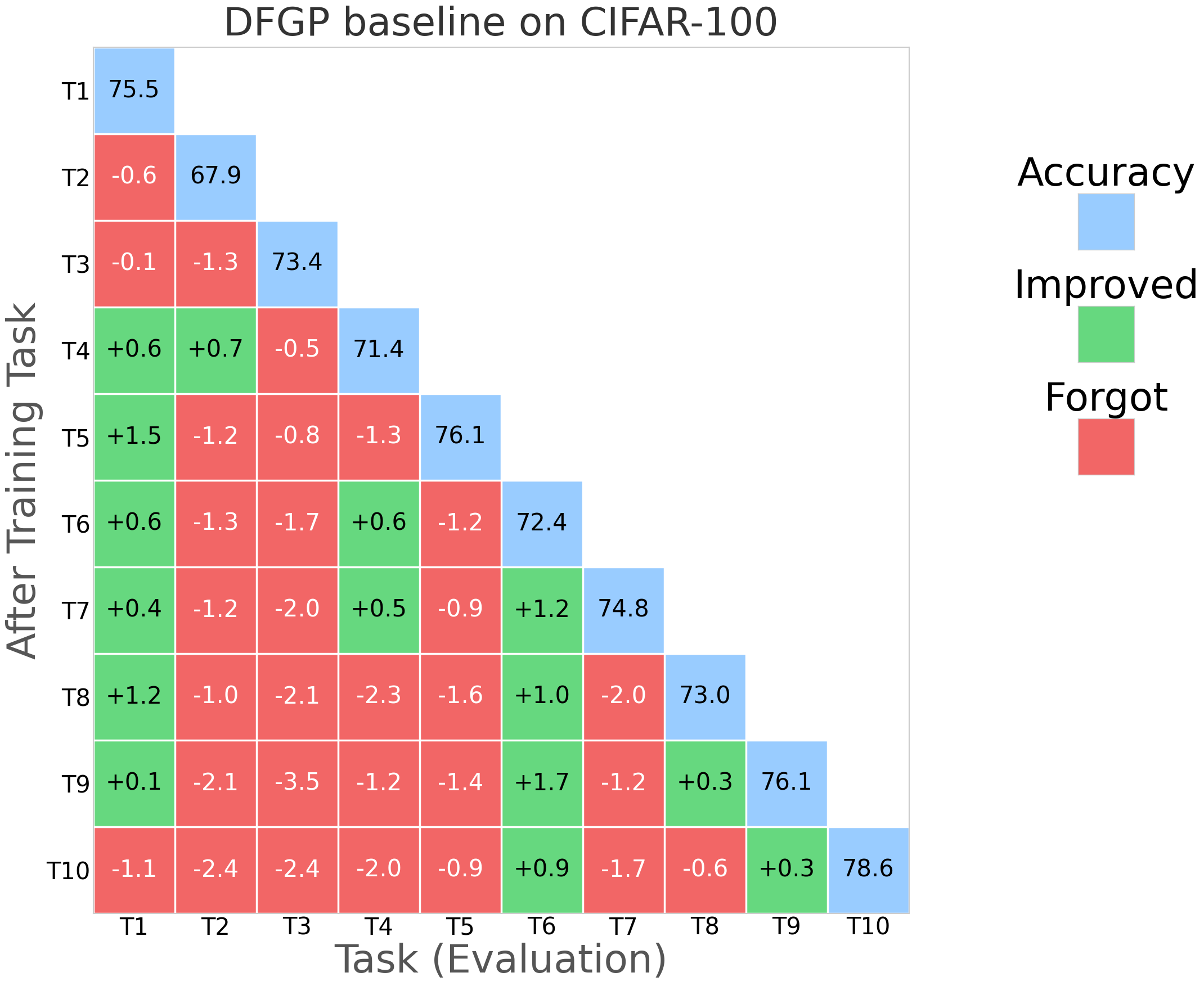}
        \caption{DFGP baseline}
    \end{subfigure}
    \caption{Performance (best setting of each method, focusing on forgetting rate, with green color denotes improving performance and red color denotes losing performance, compared to the time the task firstly trained) of (a) NESS compared to (b) SGP; (c) TRGP; (d) DFGP on CIFAR-100 experiment (seed 3).}
    \label{fig:cifar_seed3}
\end{figure}

\begin{figure}[t]
    \centering
    \begin{subfigure}[b]{0.48\textwidth}
        \centering
        \includegraphics[width=\textwidth]{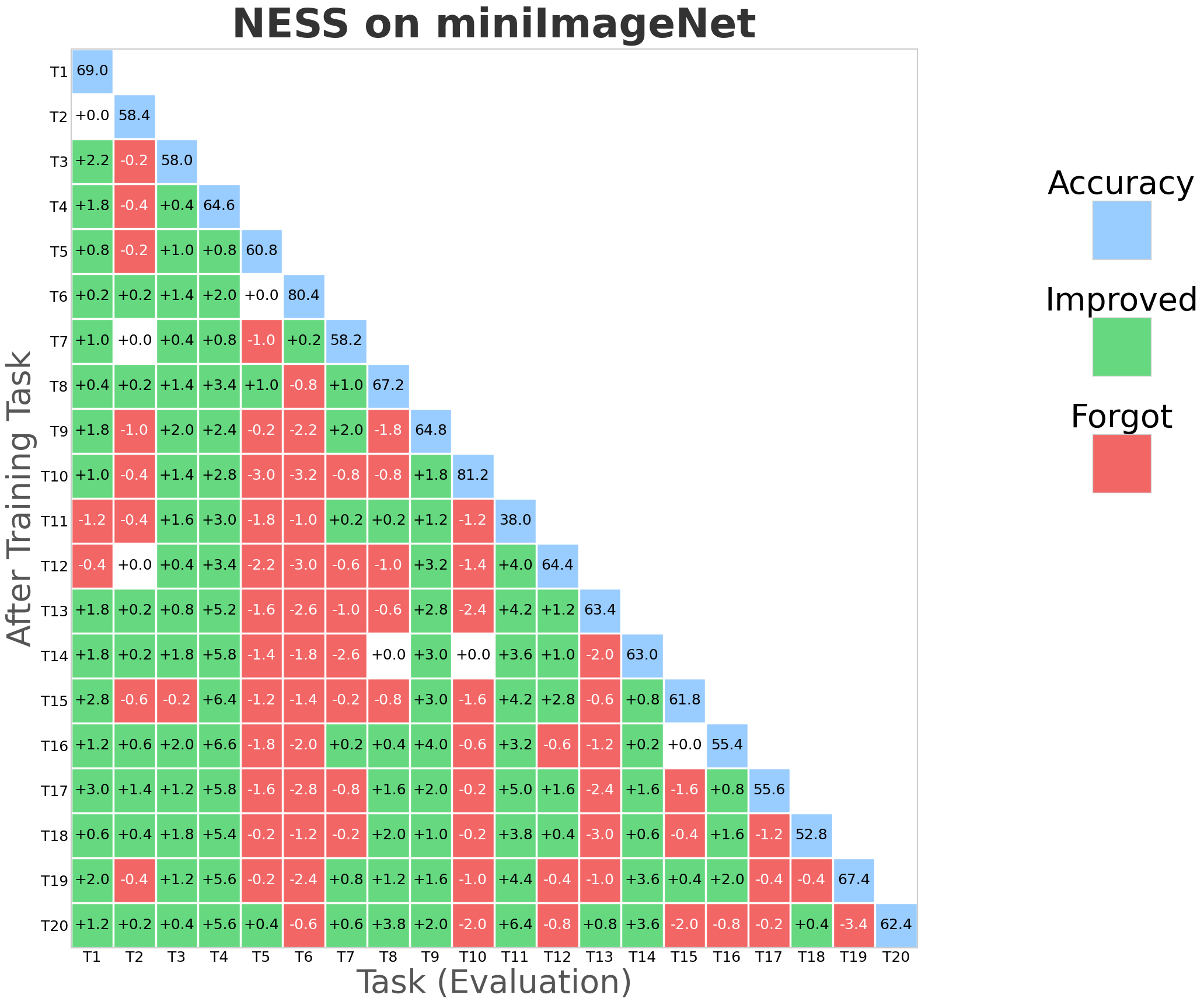}
        \caption{NESS}
    \end{subfigure}
    \hfill
    \begin{subfigure}[b]{0.48\textwidth}
        \centering
        \includegraphics[width=\textwidth]{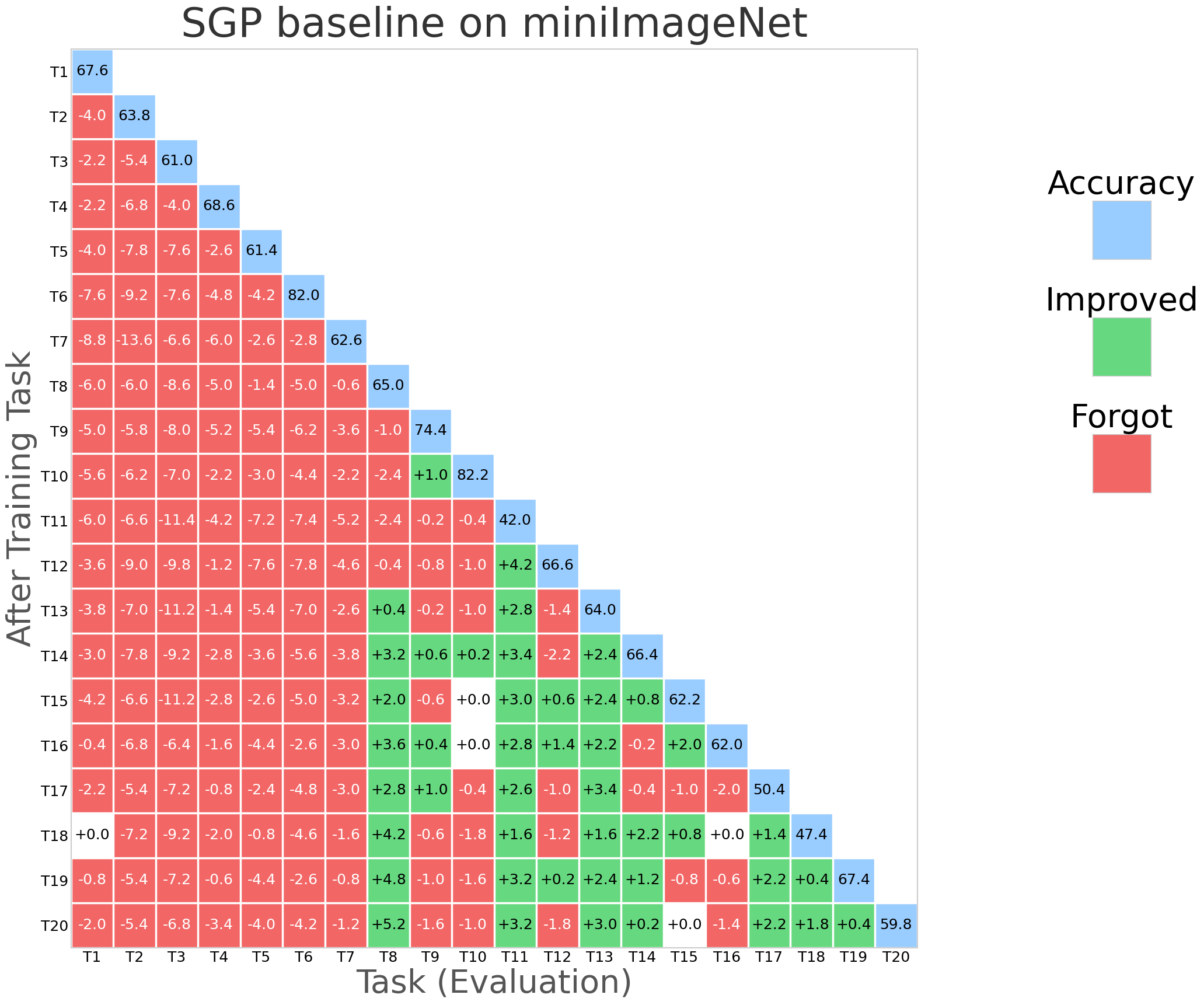}
        \caption{SGP baseline}
    \end{subfigure}
    
    \vspace{0.5cm}
    
    \begin{subfigure}[b]{0.48\textwidth}
        \centering
        \includegraphics[width=\textwidth]{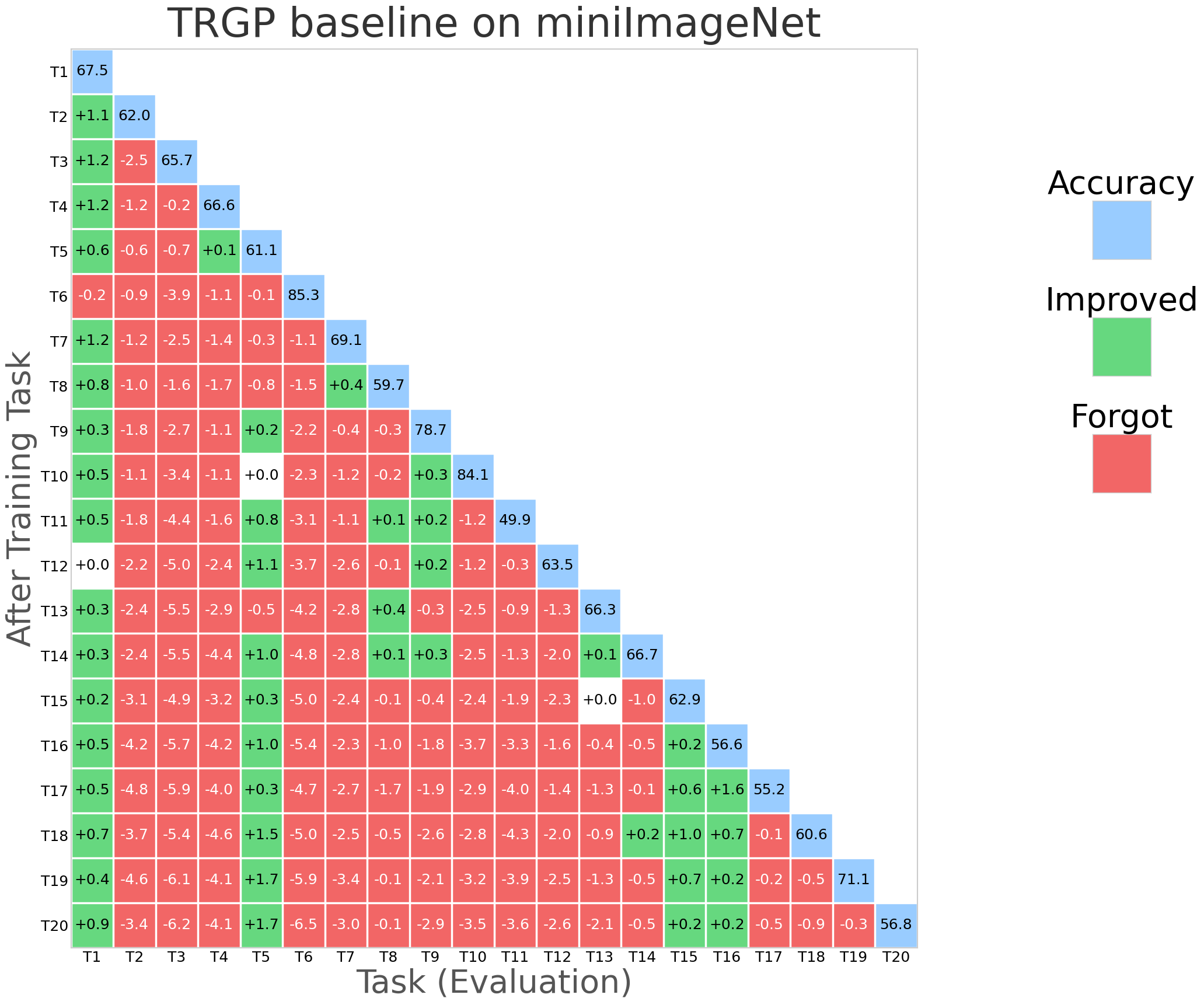}
        \caption{TRGP baseline}
    \end{subfigure}
    \hfill
    \begin{subfigure}[b]{0.48\textwidth}
        \centering
        \includegraphics[width=\textwidth]{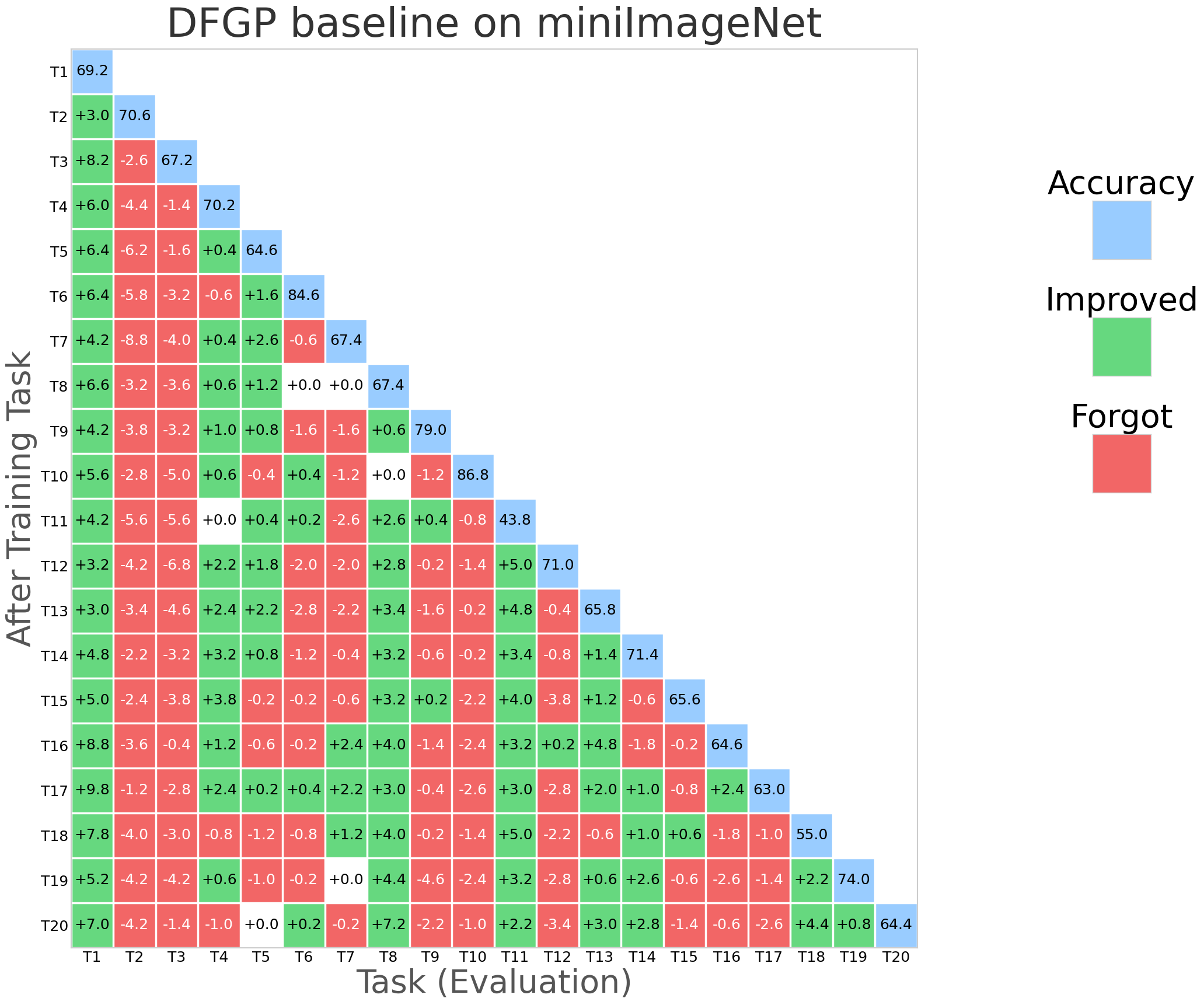}
        \caption{DFGP baseline}
    \end{subfigure}
    \caption{Performance (best setting of each method, focusing on forgetting rate, with green color denotes improving performance and red color denotes losing performance, compared to the time the task firstly trained) of (a) NESS compared to (b) SGP; (c) TRGP; (d) DFGP on miniImageNet experiment (seed 37).}
    \label{fig:mini_seed37}
\end{figure}

\textbf{Additional figures. }
Additional visualization for our method compared to the baselines is in Figure~\ref{fig:cifar_seed2}, Figure~\ref{fig:cifar_seed3} (for CIFAR-100 experiment), and in Figure~\ref{fig:mini_seed37} (for MiniImageNet experiment).

\begin{table}[ht!]
\centering
\scriptsize
\caption{Performance across 5 random seeds (detailed results) for \textbf{CIFAR-100} experiment. For SGD with momentum (SGDm), we used momentum=0.9 and we denoted $\lambda$ = weight decay.}
\vspace{0.3cm}
\begin{tabular}{|c|cc|cc|cc|cc|cc|}
\hline
 \textbf{Setting} 
    & \multicolumn{2}{c|}{\textbf{Seed 1}}
    & \multicolumn{2}{c|}{\textbf{Seed 2}}
    & \multicolumn{2}{c|}{\textbf{Seed 3}}
    & \multicolumn{2}{c|}{\textbf{Seed 4}}
    & \multicolumn{2}{c|}{\textbf{Seed 37}} \\ \cline{2-11}
&
    \textbf{ACC} & \textbf{BWT} &
    \textbf{ACC} & \textbf{BWT} &
    \textbf{ACC} & \textbf{BWT} &
    \textbf{ACC} & \textbf{BWT} &
    \textbf{ACC} & \textbf{BWT}  \\ \hline

GPM & 71.85 & -0.67 & 72.00 & -0.41 & 72.07 & -0.44 & 71.78 & 0.67 & 70.45 & -0.56 \\
SGP & \textbf{75.89} & -0.63 & \textbf{75.78} & -1.63 &\textbf{ 76.01} & -1.03 & \textbf{76.18} & -1.33 & \textbf{76.07} & -1.26  \\ 
TRGP & 75.10 & \textbf{-0.01} & 75.62 & \textbf{-0.09} & 74.92 & \textbf{-0.06} & 75.49 & \textbf{0.32} & 74.94 & \textbf{0.15} \\ 
FS-DGPM & 73.95 & -3.30  & 74.16 & -3.13 & 74.14 & -2.87 & 74.09 & -2.56 & 74.16 & -3.28  \\ 
 DFGP (lr=0.01, mixup=0.01) & 73.09 & -0.80 & 73.39 & -1.07  & 72.90 & -1.14 & 73.49 & -0.71 & 73.34 & -1.02 \\ 
 DFGP (lr=0.01, mixup=0.05) & 73.34 & -0.62 & 73.63 & -0.93 & 74.14 & -1.58 & 74.06 & -0.70 & 73.70 & -1.31  \\
DFGP (lr=0.01, mixup=0.001) & 73.09 & -0.98 & 74.11 & -0.96 & 73.45 & -1.20 & 73.15 & -0.48 & 72.80 & -1.83  \\
 DFGP (lr=0.01, mixup=0.0001) & 73.87 & -0.27 & 72.77 & -1.53 & 72.93 & -1.10 & 73.35 & -1.13 & 72.87 & -1.54 \\

DFGP (lr=0.01, no mixup) & 73.10 & -0.88 & 74.22 & -0.92 & 72.78 & -1.89 & 73.51 & -0.93 & 72.58 & -1.37  \\ \hline
\textbf{NESS (SAM, no $\lambda$, $\varepsilon_1$=0.001, lr=0.05)} & 72.25 & -0.51 & 72.61 & 0.24 & 72.57 & -1.20 & 72.17 & -0.33 & 72.97 & 0.37  \\
\textbf{NESS (SGDm, no $\lambda$, $\varepsilon_1$=0.001, lr=0.005)} & 72.09 & 0.08 & 72.14 & -0.20 & 71.86 & -0.44 & 71.59  & -0.40 & 72.48 & -0.43 \\ \hline 
\textbf{NESS (SAM, $\lambda$=0.0001, $\varepsilon_1$=0.001, lr=0.05)} & 72.47 & -0.37 & \textbf{72.67} & -0.21 & \textbf{72.58} & -0.48 & \textbf{72.53}	 & \textbf{0.71} & \textbf{73.16} & \textbf{-0.06} \\
 
\textbf{NESS (SGDm, $\lambda$=0.0001, $\varepsilon_1$=0.001, lr=0.005)} & \textbf{72.76} & \textbf{0.09} & 72.59 & \textbf{0.17} & 72.55 & \textbf{0.32} & 72.29 & 0.23 & 72.10 & -0.67 \\ \hline

\end{tabular}
\label{tab:seeds_CIFAR100}
\end{table}

\begin{table}[ht!]
\centering
\caption{Performance across 5 random seeds (average performance) for \textbf{CIFAR-100} experiment. For SGD with momentum (SGDm), we used momentum=0.9 and we denoted $\lambda$ = weight decay.}
\vspace{0.3cm}
\begin{tabular}{|c|cc|}
\hline
 \textbf{Setting} & \multicolumn{2}{c|}{\textbf{Mean $\pm$ std}}\\ \cline{2-3}
&
    \textbf{ACC} & \textbf{BWT} \\ \hline

GPM & 71.63 $\pm$ 0.67 & -0.28 $\pm$ 0.54\\
SGP & \textbf{75.99 $\pm$ 0.16} & -1.18 $\pm$ 0.37 \\ 
TRGP & 75.21 $\pm$ 0.32 &  \textbf{0.06 $\pm$ 0.17} \\ 
FS-DGPM & 74.10 $\pm$ 0.09 & -3.03 $\pm$ 0.31\\ 
DFGP (lr=0.01, mixup=0.01) & 73.24 $\pm$ 0.24 & -0.95 $\pm$ 0.18\\ 
DFGP (lr=0.01, mixup=0.05) & 73.77 $\pm$ 0.33 & -1.03 $\pm$ 0.41 \\
DFGP (lr=0.01, mixup=0.001) & 73.32 $\pm$ 0.50 & -1.09 $\pm$ 0.49 \\
DFGP (lr=0.01, mixup=0.0001) & 73.16 $\pm$ 0.46 & -1.11 $\pm$ 0.52 \\

DFGP (lr=0.01, no mixup) & 73.24 $\pm$ 0.65 & -1.20 $\pm$ 0.44 \\ \hline
\textbf{NESS (SAM, no $\lambda$, $\varepsilon_1$=0.001, lr=0.05)} & 72.43 $\pm$ 0.21 & -0.46 $\pm$ 0.51 \\
 
\textbf{NESS (SGDm, no $\lambda$, $\varepsilon_1$=0.001, lr=0.005)} & 72.03 $\pm$ 0.33 & -0.28 $\pm$ 0.22 \\ \hline
\textbf{NESS (SAM, $\lambda$=0.0001, $\varepsilon_1$=0.001, lr=0.05)} & \textbf{72.56 $\pm$ 0.07} & -0.17 $\pm$ 0.51 \\
 
\textbf{NESS (SGDm, $\lambda$=0.0001, $\varepsilon_1$=0.001, lr=0.005)} & 72.46 $\pm$ 0.26 & \textbf{0.03 $\pm$ 0.40} \\ \hline

\end{tabular}
\label{tab:seeds_avg_CIFAR100}
\end{table}

\begin{table}[ht!]
\centering
\scriptsize
\caption{Performance across 5 random seeds (detailed results) for \textbf{5-datasets} experiment. For SGD with momentum (SGDm), we used momentum=0.9 and we denoted $\lambda$ = weight decay.}
\vspace{0.3cm}
\begin{tabular}{|c|cc|cc|cc|cc|cc|}
\hline
\textbf{Setting} 
    & \multicolumn{2}{c|}{\textbf{Seed 1}}
    & \multicolumn{2}{c|}{\textbf{Seed 2}}
    & \multicolumn{2}{c|}{\textbf{Seed 3}}
    & \multicolumn{2}{c|}{\textbf{Seed 4}}
    & \multicolumn{2}{c|}{\textbf{Seed 37}} \\ \cline{2-11}
&
    \textbf{ACC} & \textbf{BWT} &
    \textbf{ACC} & \textbf{BWT} &
    \textbf{ACC} & \textbf{BWT} &
    \textbf{ACC} & \textbf{BWT} &
    \textbf{ACC} & \textbf{BWT} \\ \hline

GPM & 91.27 & -0.81 & 89.98 & -1.52 & 91.1 & -1.08 & 90.12 & -1.14 & 90.56 & -1.31 \\
SGP & 91.20 & -1.70 & 90.16 & -1.88 & 90.72 & -1.98 & 90.03 & -1.88 & 90.30 & -1.67 \\ 
TRGP & \textbf{93.59} & \textbf{0.03} & \textbf{92.24} & \textbf{-0.12} & \textbf{93.32} & \textbf{-0.08} & \textbf{92.12} & \textbf{-0.18}  & \textbf{92.61} & \textbf{-0.12}  \\ 
DFGP (lr=0.1, mixup=0.01) & 91.12 & -2.67 & 91.48 & -2.47 & 91.48 & -2.22 & 91.55 & -2.15 &  91.72 & -1.86 \\ 
DFGP (lr=0.1, mixup=0.05) & 89.84 & -4.54 & 90.70 & -3.28 & 89.65 & -4.30 & 90.34 & -3.72 & 90.57 & -3.50 \\ 
DFGP (lr=0.1, mixup=0.001) & \textbf{91.92} & \textbf{-1.60} & 91.93 & -1.87 & 91.04 & -2.21 & 91.04 & -1.99 & 92.02 & -1.64 \\
DFGP (lr=0.1, mixup=0.0001) & 91.55 & -1.85 & 92.10  & -1.73 & 92.05 & -1.54 & 90.81 & -2.51 & 91.05 & -2.35 \\

DFGP (lr=0.1, no mixup) & 91.53 & -2.16 & 91.90 & -1.66 & 91.71 & -1.59 & 91.71 & -1.55 & 91.53 & -1.54 \\ \hline
\textbf{NESS (SAM, no $\lambda$, $\varepsilon_1$=0.001, lr=0.1)} & 90.61 & -0.76 & 90.71 & -0.98 & 90.86 & -0.70 & 90.87 & -0.74 & 90.45 & -0.98 \\
\textbf{NESS (SGDm, no $\lambda$, $\varepsilon_1$=0.001, lr=0.05)} & 89.57 & -0.56 & 90.26 & -0.66 & 90.44 & -0.74 & 90.35 & -0.53 & 89.79 & \textbf{-0.49} \\ 
\textbf{NESS (SGDm, no $\lambda$, $\varepsilon_1$=0.001, lr=0.01)} & 90.32 & -1.15 & 89.87 & -0.81 & 90.58 & -0.87 & 90.19 & -1.09 & 90.69 & -1.00 \\ \hline
\textbf{NESS (SAM, $\lambda$=0.00005, $\varepsilon_1$=0.001, lr=0.1)} & \textbf{90.90} & -1.01 & \textbf{91.08} & -1.06 & \textbf{91.00} & -1.10 & \textbf{90.99} & \textbf{-0.45} & \textbf{90.91} & \textbf{-0.68} \\

\textbf{NESS (SGDm, $\lambda$=0.00005, $\varepsilon_1$=0.001, lr=0.05)} & 90.57 & \textbf{-0.36} & 90.73 & \textbf{-0.59} & 90.25 & \textbf{-0.59} & 89.79 & -0.58 & 89.66 & -0.79  \\ 
\hline

\end{tabular}
\label{tab:seeds_FIVE}
\end{table}

\begin{table}[ht!]
\centering
\caption{Performance across 5 random seeds (average performance) for \textbf{5-datasets} experiment. For SGD with momentum, we used momentum=0.9 and we denoted $\lambda$ = weight decay.}
\vspace{0.3cm}
\begin{tabular}{|c|cc|}
\hline
 \textbf{Setting} & \multicolumn{2}{c|}{\textbf{Mean $\pm$ std}}\\ \cline{2-3}
&
    \textbf{ACC} & \textbf{BWT} \\ \hline

GPM & 90.61 $\pm$ 0.57 & -1.17 $\pm$ 0.26 \\
SGP & 90.48 $\pm$ 0.48 & -1.82 $\pm$ 0.13 \\ 
TRGP & \textbf{92.78 $\pm$ 0.65} & \textbf{-0.09 $\pm$ 0.08}  \\ 
DFGP (lr=0.1, mixup=0.01) & 91.47 $\pm$ 0.22 & -2.27 $\pm$ 0.31 \\ 
DFGP (lr=0.1, mixup=0.05) & 90.22 $\pm$ 0.46 & -3.87 $\pm$ 0.53 \\
DFGP (lr=0.1, mixup=0.001) & 91.59 $\pm$ 0.50 & -1.86 $\pm$ 0.25 \\
DFGP (lr=0.1, mixup=0.0001) & 91.51 $\pm$ 0.15 & -1.70 $\pm$ 0.42 \\ 

 DFGP (lr=0.1, no mixup) & 91.68 $\pm$ 0.15 & -1.70 $\pm$ 0.26 \\ \hline
 
\textbf{NESS (SAM, no $\lambda$, $\varepsilon_1$=0.001, lr=0.1)} & 90.70 $\pm$ 0.18 & -0.83 $\pm$ 0.14\\

\textbf{NESS (SGDm, no $\lambda$, $\varepsilon_1$=0.001, lr=0.05)} & 90.08 $\pm$ 0.38 & -0.60 $\pm$ 0.10 \\ 
\textbf{NESS (SGDm, no $\lambda$, $\varepsilon_1$=0.001, lr=0.01)} & 90.33 $\pm$ 0.33 & -0.98 $\pm$ 0.14 \\ 
\hline
\textbf{NESS (SAM, $\lambda$=0.00005, $\varepsilon_1$=0.001, lr=0.1)} & \textbf{90.98 $\pm$ 0.07} & -0.86 $\pm$ 0.28 \\
\textbf{NESS (SGDm, $\lambda$=0.00005, $\varepsilon_1$=0.001, lr=0.05)} & 90.20 $\pm$ 0.47 & \textbf{-0.58 $\pm$ 0.15} \\ 
\hline

\end{tabular}
\label{tab:seeds_avg_FIVE}
\end{table}

\begin{table}[ht!]
\centering
\scriptsize
\caption{Performance across 5 random seeds (detailed result) for \textbf{MiniImageNet} experiment. For SGD with momentum (SGDm), we used momentum=0.9 and we denoted $\lambda$ = weight decay.}
\vspace{0.3cm}
\begin{tabular}{|c|cc|cc|cc|cc|cc|}
\hline
\textbf{Setting} 
    & \multicolumn{2}{c|}{\textbf{Seed 1}}
    & \multicolumn{2}{c|}{\textbf{Seed 2}}
    & \multicolumn{2}{c|}{\textbf{Seed 3}}
    & \multicolumn{2}{c|}{\textbf{Seed 4}}
    & \multicolumn{2}{c|}{\textbf{Seed 37}} \\ \cline{2-11}
&
    \textbf{ACC} & \textbf{BWT} &
    \textbf{ACC} & \textbf{BWT} &
    \textbf{ACC} & \textbf{BWT} &
    \textbf{ACC} & \textbf{BWT} &
    \textbf{ACC} & \textbf{BWT} \\ \hline

GPM & 61.14 & -3.54	 & 66.59 & -0.41 & 61.52 & -1.95 & 65.53 & -0.82 & 63.03 & -0.25 \\
SGP & 65.42 & -0.96 & 66.71 & -0.14 & 61.41 & -1.16 & 65.71 & \textbf{0.48} & 63.00 & -0.88 \\ 
TRGP & 61.70 & -1.34 & 65.06 & -0.38 & 59.59 & -1.99 & 63.74 & -0.50 & 63.61 & -1.96 \\ 
DFGP (lr=0.1, mixup=0.1) & 67.18 & -0.33 & 68.82 & -0.53 & 68.25 & -0.87 & \textbf{69.53}  & -0.74 & \textbf{68.76} & \textbf{0.51} \\ 
DFGP (lr=0.1, mixup=0.05) & 68.55 & 0.58 & 70.49 & \textbf{1.56} & 66.25 & -1.79 & 68.63 & -1.20 & 68.56 & 0.28 \\
DFGP (lr=0.1, mixup=0.01) & \textbf{70.18} & 1.29 & \textbf{71.42} & 0.99 & 66.12 & -2.60 & 68.78 & -0.74 & 66.68 & -1.09 \\
DFGP (lr=0.1, mixup=0.001) & 67.19 & -0.84 & 70.56 & 1.02 & 66.66 & -3.20 & 68.41 & -1.17 & 65.93 & -1.35 \\
DFGP (lr=0.1, mixup=0.0001) & 69.40 & \textbf{1.59} & 68.30 & 0.43 & 67.97 & -1.78 & 69.41 & 0.43 & 66.87 & -1.45 \\ 

DFGP (lr=0.1, no mixup) & 67.13 & 0.33 & 69.89 & 0.48 & 67.16 & -2.84 & 68.66 & 0.21 & 65.52 & -0.28 \\ \hline
\textbf{NESS (SAM, no $\lambda$, $\varepsilon_1$=0.0005, lr=0.1)} & 63.35 & \textbf{0.77} & 62.2 & -0.72 & 63.32 & -1.75 & 63.55 & \textbf{0.18} & 59.53 & -0.83 \\

\textbf{NESS (SGDm, no $\lambda$, $\varepsilon_1$=0.0005, lr=0.01)} & 62.80 & 0.60 & \textbf{66.27} & \textbf{1.55} & 62.57 & -1.63 & 62.11 &   -0.39 & \textbf{64.40} & \textbf{1.41} \\  \hline

\textbf{NESS (SAM, $\lambda$=0.00001, $\varepsilon_1$=0.0005, lr=0.1)} & 61.32 & -0.95 & 63.37 & -0.44 & 63.55 & 0.39 & 65.07 & -0.79 & 64.11 & 0.49  \\
\textbf{NESS (SGDm, $\lambda$=0.00001, $\varepsilon_1$=0.0005, lr=0.01)} &\textbf{63.69} & 0.64 & 64.29 & 0.82 & \textbf{63.47} & \textbf{0.36} & \textbf{64.02} & -0.57 & 63.12 & 0.82 \\  \hline

\end{tabular}
\label{tab:seeds_MINI}
\end{table}

\begin{table}[ht!]
\centering
\caption{Performance across 5 random seeds (average performance) for \textbf{MiniImageNet} experiment. For SGD with momentum, we used momentum=0.9 and we denoted $\lambda$ = weight decay.}
\vspace{0.3cm}
\begin{tabular}{|c|cc|}
\hline
 \textbf{Setting} & \multicolumn{2}{c|}{\textbf{Mean $\pm$ std}}\\ \cline{2-3}
&
    \textbf{ACC} & \textbf{BWT} \\ \hline

GPM & 63.56 $\pm$ 2.42 & -1.39 $\pm$ 1.37 \\
SGP & 64.45 $\pm$ 2.18 & -0.53 $\pm$ 0.68 \\ 
TRGP & 62.74 $\pm$ 2.13 & -1.23 $\pm$  0.77 \\ 
DFGP (lr=0.1, mixup=0.1) & 68.51 $\pm$ 0.87 & -0.39 $\pm$ 0.54 \\ 
DFGP (lr=0.1, mixup=0.05) & 68.50 $\pm$ 1.50 & \textbf{-0.11  $\pm$ 1.36} \\
DFGP (lr=0.1, mixup=0.01) & \textbf{68.64 $\pm$ 2.25} & -0.43 $\pm$ 1.60 \\
DFGP (lr=0.1, mixup=0.001) & 67.75 $\pm$ 1.81 & -1.11 $\pm$ 1.50 \\
DFGP (lr=0.1, mixup=0.0001) & 68.39 $\pm$ 1.07 & -0.16 $\pm$ 1.42 \\ 

DFGP (lr=0.1, no mixup) & 67.67 $\pm$ 1.66 &  -0.42 $\pm$ 1.38  \\ \hline
\textbf{NESS (SAM, no $\lambda$, $\varepsilon_1$=0.0005, lr=0.1)} & 62.39 $\pm$ 1.68 & -0.47 $\pm$ 0.97  \\
\textbf{NESS (SGDm, no $\lambda$, $\varepsilon_1$=0.0005, lr=0.01)} & 63.63 $\pm$ 1.71 & 0.31 $\pm$ 1.33 \\  \hline
\textbf{NESS (SAM, $\lambda$=0.00001, $\varepsilon_1$=0.0005, lr=0.1)} & 63.48 $\pm$ 1.38 & -0.26 $\pm$ 0.67  \\
\textbf{NESS (SGDm, $\lambda$=0.00001, $\varepsilon_1$=0.0005, lr=0.01)} & \textbf{63.72 $\pm$ 0.46} & \textbf{0.41 $\pm$ 0.58} \\  \hline

\end{tabular}
\label{tab:seeds_avg_MINI}
\end{table}

\subsection{Limitations and Future Work}

\textbf{Effect of threshold $\varepsilon_1$. }
Although NESS achieves promising results in reducing catastrophic forgetting, we observe that with a high threshold $\varepsilon_1$, the model approaches unconstrained continual learning on each task, leading to a very high forgetting rate. Moreover, using a threshold $\varepsilon_1$ that is too small, our model may struggle to learn new tasks due to overly restrictive constraints. Therefore, tuning $\varepsilon_1$ remains an open problem.
In addition, because every layer has its own shape, we also observed that some layers may have nearly full-rank trainable matrices. Although this does not significantly affect overall performance and we can tolerate a slight drop in performance or a slightly higher forgetting rate, we still consider designing a more effective threshold $\varepsilon_1$ selection strategy as a challenge for future research.
Finally, because NESS requires input from previous tasks (though not for retraining), it may be unrealistic in many real-world settings. Therefore, developing an approach with limited access to input remains an open problem.

\textbf{Effect of Biases and BatchNorm Layers. }
In our experiments, we only process linear/convolutional layers due to their significant impact on model performance. We do not apply NESS to biases and BatchNorm layers, which may cause the model to lose some capacity and result in a slight performance loss. However, we believe that applying NESS to these components requires a different technique, so we leave it as an open problem for future work.

\end{document}